\documentclass[lettersize,journal]{IEEEtran}
\usepackage{amsmath,amsfonts}
\usepackage{algorithmic}
\usepackage{algorithm}
\usepackage{array}
\usepackage[caption=false,font=normalsize,labelfont=sf,textfont=sf]{subfig}
\usepackage{textcomp}
\usepackage{stfloats}
\usepackage{url}
\usepackage{verbatim}
\usepackage{graphicx}
\usepackage{cite}

\usepackage{booktabs}
\usepackage{multirow}
\usepackage{colortbl}
\usepackage{amsmath}
\usepackage{xcolor}
\usepackage{amssymb}
\usepackage{pifont}
\usepackage{makecell}

\usepackage[
    colorlinks=true,       
    linkcolor=blue,        
    citecolor=blue,         
    urlcolor=black,        
]{hyperref}

\newif\ifrevisedtwo
\revisedtwofalse   

\newcommand{\revtwo}[1]{%
  \ifrevisedtwo
    \textcolor{red}{#1}%
  \else
    #1%
  \fi
}

\newif\ifrevised
\revisedfalse   

\newcommand{\rev}[1]{%
  \ifrevised
    \textcolor{red}{#1}%
  \else
    #1%
  \fi
}

\begin{document}

\title{SRENet: Spectral Re-Entry Network for\\ Point Cloud Action Recognition}

\author{Qiuxia Wu,~\IEEEmembership{Member,~IEEE,}
Jiarui Lan,
Wenxiong Kang,~\IEEEmembership{Member,~IEEE,}
Zhiyong Wang,~\IEEEmembership{Member,~IEEE,} \\
Kun Hu,~\IEEEmembership{Member,~IEEE}
\thanks{Qiuxia Wu and Jiarui Lan are with the School of Software Engineering, South China University of Technology, Guangzhou 511400, China. (e-mail: qxwu@scut.edu.cn, jiaruilan@foxmail.com)}
\thanks{Wenxiong Kang is with the School of Automation Science and Engineering, South China University of Technology, Guangzhou 511400, China. (e-mail: auwxkang@scut.edu.cn)}
\thanks{Zhiyong Wang is with the School of Computer Science, University of Sydney, Sydney, NSW 2006, Australia. (e-mail: zhiyong.wang@sydney.edu.au)}
\thanks{Kun Hu is with the School of Science, Edith Cowan University, Perth, WA 6027, Australia. (e-mail: hukun\_sdu@hotmail.com)}
\thanks{This work is supported by the ECU Early-Mid Career Researcher (EMCR) Grant Scheme.}
}


\IEEEpubid{\begin{minipage}{\textwidth}\ \centering
        Copyright~\copyright~2026 IEEE. Personal use of this material is permitted. \\
        However, permission to use this material for any other purposes must be obtained from the IEEE by sending an email to pubs-permissions@ieee.org.
\end{minipage}}

\maketitle

\begin{abstract}
Recognizing human actions from point cloud sequences is critical for 3D perception driven applications such as autonomous driving and human-computer interaction.  
However, the irregular structure and temporal inconsistency of point clouds pose unique challenges for spatio-temporal representation learning, especially in capturing both global motion context and fine-grained temporal dynamics.
We propose \textbf{SRENet}, a spectral-aware framework designed to explicitly learn both global context and fine-grained temporal dynamics of motion from a frequency perspective for action recognition. 
SRENet introduces a \textit{Spectral Decomposition Block} (SDeBlock) that performs wavelet-based analysis along temporal and spatial axes, disentangling features into low- and high-frequency components with frequency-specific attention.  
To recover residual dynamics and re-align temporal frequency structures distorted during semantic fusion, a \textit{Spectral Re-entry Block} (SReBlock) performs secondary temporal decomposition.  
Furthermore, a spectral-aware learning strategy is devised to enhance discriminability in both frequency subspaces via contrastive loss and a curriculum schedule that gradually shifts focus from low- to high-frequency spaces in line with coarse to detailed motion patterns.
Extensive experiments on MSR-Action3D, NTU-RGBD and NTU-RGBD120 demonstrate that SRENet achieves state-of-the-art performance, validating the effectiveness of frequency modeling in point cloud-based action understanding.
\footnote{
The source code is available at https://github.com/tomlan2026/SRENet.
}
\end{abstract}


\section{Introduction}
\label{sec:intro}

\IEEEPARstart{R}{ecognizing} human actions from point cloud sequences is essential for 3D perception in various scenarios such as autonomous driving, surveillance, and human-computer interaction. 
While point clouds provide accurate geometric information and strong robustness to lighting and background variations~\cite{fan2021p4transformer}, their irregular structure and temporal misalignment make spatio-temporal modeling particularly challenging~\cite{fan2022pst-transformer}. 
These characteristics distinguish point cloud-based action recognition from its RGB-based counterparts and motivate the need for tailored architectures.

\begin{figure}[t]
  \centering
\includegraphics[width=0.985\linewidth]{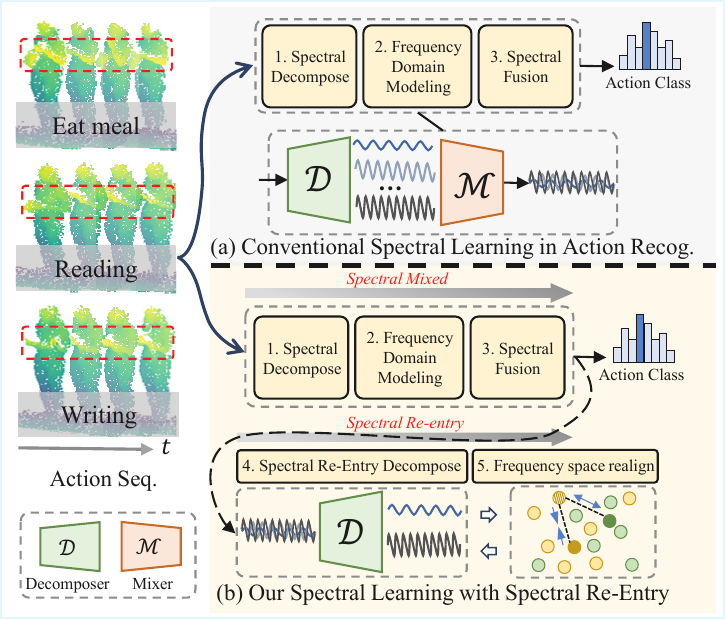}
   \caption{
   (a) Previous methods extract spectral features based on a single frequency decomposition.
(b) Our decomposition and re-entry strategy realigns spectral structures after cross-axis fusion, recovering residual temporal patterns critical for action recognition.
   }
   \label{fig:intro}
\end{figure}


To address these challenges, recent work has explored Transformer-like architectures~\cite{fan2021p4transformer, fan2022pst-transformer, wen2022pptr, wei2022pst2} to model long-range spatio-temporal dependencies in point cloud sequences.
While multi-head self-attention enables global context modeling, it has been shown to exhibit low-pass filtering behavior in vision tasks~\cite{park2022how-vit-work}, suppressing high-frequency signals and emphasizing coarse structures. 
This low-frequency bias compromise the model’s ability to capture fine-grained motion cues—crucial for distinguishing subtle actions such as \textit{eating}, \textit{reading}, and \textit{writing} (Fig.~\ref{fig:intro}).
In point cloud sequences, low-frequency components reflect the overall motion trajectory and structure, while high-frequency components encode localized, fine-grained motion dynamics. 
Therefore, effective action recognition requires a representation that integrates both low- and high-frequency information.
\IEEEpubidadjcol

\begin{figure}[t]
  \centering
\includegraphics[width=0.985\linewidth]{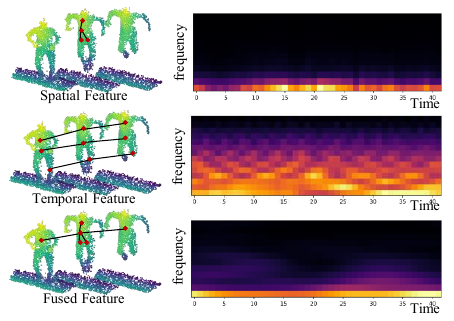}
   \caption{
   Time-frequency amplitude map of the action ‘\textit{jump up}’. 
   The vertical axis represents frequency from low to high and the horizontal denotes time. Brighter colors indicate larger amplitudes.
   }
   \label{fig:intro-vis}
\end{figure}

Inspired by the success of frequency-based modeling in well-structured vision and time-series tasks~\cite{pan2022hilo-vit, si2022inception-transformer, wu2024freqmixformer, gao2023freq-motion}, we explore how frequency information can benefit motion understanding in point cloud sequences. 
Prior approaches typically decompose input signals into multiple frequency bands, process each band separately, and fuse the outputs to capture multi-scale semantics (Fig.~\ref{fig:intro}a). 
However, directly applying such strategies to point clouds is non-trivial due to their unordered structure and temporal inconsistency. In particular, inter-frame point misalignment disrupts the temporal coherence required for reliable frequency decomposition, making it difficult to model frequency-specific patterns.
Moreover, simply fusing frequency components—especially across spatial and temporal dimensions—may weaken certain underlying frequency structures, as illustrated in Figure \ref{fig:intro-vis}.
Spatial features are concentrated in low-frequency narrow bands; in contrast, temporal features cover wideband regions that incorporate high-frequency ranges. After cross-axis fusion, a novel structure distinct from both is generated, which preserves temporal continuity while weakening temporal high-frequency features.
This frequency weakening poses greater challenges to preserve the spectral patterns that are most informative for action discrimination, thus making it necessary to restore clear frequency structures.

To address these challenges, we propose the {Spectral Re-Entry Network (SRENet)}, a spectrum-aware framework tailored for action recognition in point cloud sequences. 
SRENet comprises two key components: a {Spectral Decomposition Block} (SDeBlock) and a {Spectral Re-entry Block} (SReBlock), and a spectral-aware learning strategy.
SDeBlock performs wavelet-based decomposition along both spatial and temporal axes to disentangle the input features into high- and low-frequency components. 
Specifically, a Spectral Temporal Attention (STA) module processes temporal frequency bands, while a  Spectral Spatial Attention (SSA) module handles spatial frequency bands.
The outputs from SSA and STA are then fused to form a frequency-aware representation for subsequent processing. 
SReBlock performs secondary wavelet decomposition to realign temporal-frequency structures disrupted inevitably during SDeBlock fusion, enhancing the temporal dynamics for further action discrimination. 
Spectral-aware learning achieves a flexible strategy for the optimization from both low- and high-frequency patterns: a contrastive loss separately enhances discriminability in low-/high-frequency subspaces, while a curriculum schedule dynamically shifts modeling focus from structural (low-frequency) to detailed (high-frequency) patterns during training. 
Comprehensive experiments on three widely-used benchmark datasets - MSR-Action3D \cite{li2010msr-action3d}, NTU-RGBD \cite{shahroudy2016ntu-rgbd} and NTU-RGBD120 \cite{liu2019ntu120} demonstrate our state-of-the-art performance. 

In summary, our key contributions are as follows:
\begin{itemize}
  \item We propose {SRENet}, a novel spectrum-aware framework for action recognition in point cloud sequences. To the best of our knowledge, this is the first work to introduce frequency analysis to this field.
  
  \item We design SDeBlock to disentangle spatial and temporal features into high- and low-frequency components with frequency-specific attention.

  \item We design SReBlock to re-align temporal-frequency structures disrupted, enhancing discriminative learning of fine-grained motion dynamics.

  \item We develop a {spectral-aware learning strategy} to enhance discriminability in both frequency subspaces via contrastive learning and a frequency-aware curriculum to guide training from coarse-to-fine motion patterns.

  \item Extensive experiments on three benchmarks demonstrate the state-of-the-art performance of our SRENet.
\end{itemize}


\section{Related Work}
\label{sec:related}

\subsection{Static Point Cloud Understanding}
3D point cloud understanding has primarily relied on \textit{spatial domain analysis}, using geometric features to characterize static scenes.
Early deep learning approaches directly operated on raw point sets, exemplified by PointNet~\cite{qi2017pointnet} and its hierarchical extension PointNet++~\cite{qi2017pointnet++}, which learn permutation-invariant point-wise features and local geometric structures.
Subsequent methods further enhanced spatial feature modeling through graph-based representations~\cite{wang2019dgcnn} and attention mechanisms~\cite{guo2021pct, wu2024Point-transformer-v3}. 
These advances have driven progress in shape classification~\cite{ma2022rethinking, su2025ri}, object detection~\cite{zhang2024safdnet, zhang2024voxel-mamba}, scene segmentation~\cite{zhu2024no-time-to-train, zhao2024unimix} and point cloud completion \cite{wu2025dc, xiao2023unsupervised-pc-completion}. 
While effective for spatial modeling, these methods operate purely in the spatial domain and lack the ability to model temporal dynamics in point cloud sequences.

\vspace{-6pt}
\subsection{Sequential Point Cloud Understanding}
Building on static point cloud representations, sequence understanding aims to capture spatio-temporal dynamics for modeling motion patterns. 
Early approaches converted point clouds into voxels or other regular structures to enable grid-based convolutions~\cite{wang20203dv, wang2024semantic-complete-4d, tong2024one-shot-pc-action}. 
 Subsequent approaches directly operate on raw point clouds, employing CNN-based architectures~\cite{fan2021pstnet, fan2021pstnet2}, MLP-based~\cite{li2022SequentialPointNet, liu2021geometrymotion}, Transformer-based~\cite{wen2022pptr, wei2022pst2, fan2021p4transformer, fan2022pst-transformer}, \revtwo{semi-supervised~\cite{chen2022maple} or }self-supervised~\cite{sheng2023pointcpsc, shen2023pointcmp} models to learn spatio-temporal relationships.

\begin{figure*}[t]
    \centering
\includegraphics[width=0.985\linewidth]{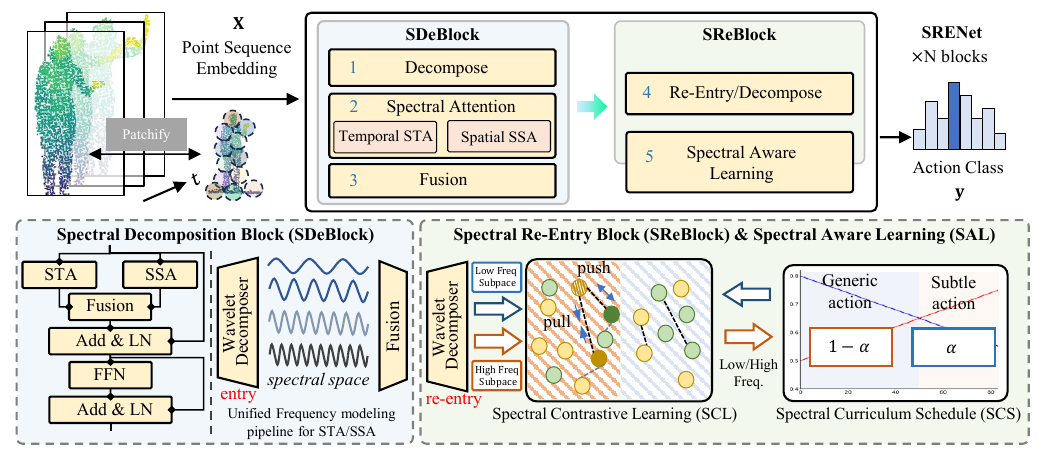}
   \caption{
Overview of SRENet, which integrates spectral decomposition and re-entry blocks for point cloud action recognition. SDeBlock extracts spatial and temporal spectral features, which are fused and re-aligned by SReBlock.
Spectral-aware learning combines contrastive loss and curriculum scheduling in low- and high-subspaces for coarse-to-fine motion learning.
   }
    \label{fig:overview}
\end{figure*}

Notably, Transformer~\cite{vaswani2017transformer} and Mamba~\cite{gu2023mamba} architectures further incorporate intrinsic part-whole relationships.
For instance, PST-Transformer \cite{fan2022pst-transformer} explicitly decouples spatial and temporal feature encoding with a Transformer; Mamba4D \cite{liu2024mamba4d} introduces a state-space model-based architecture \cite{gu2023mamba}, enhancing the efficiency for long-range spatio-temporal sequence modeling. 
While existing approaches have advanced spatio-temporal modeling in point cloud sequences, frequency-domain characteristics remain largely underexplored.

\vspace{-6pt}
\subsection{Frequency Representation Learning}
Frequency representation learning applies spectral transforms to extract frequency-domain features that enhance feature discriminability. 
This approach has demonstrated significant effectiveness across a variety of domains, such as image compression \cite{wang2024fdnet}, face forgery detection \cite{zhang2024freq-face-forgery-detection} and image classification~\cite{song2024freq-image-cls}.
In the forgery detection domain, recent works further reveal that discriminative knowledge can be encoded in latent units \cite{dou2026dna} or shared across modalities \cite{dou2026beyond}.
\revtwo{Recent studies~\cite{yang2022delving-freq, wu2024freqmixformer, chang2024skele-wave-contrastive, qin2022strengthening} have explored frequency-based modeling for human motion understanding in structured modalities such as skeleton sequences or video frames, where joint positions are fixed and follow consistent topological order.} 
This structural regularity facilitates the use of spectral transforms. 
Yet, such assumptions do not hold in point cloud sequences, where points are unordered and spatially inconsistent across frames, posing significant challenges for frequency-domain analysis.

Although spectral methods have been applied to static point clouds~\cite{shuman2013graph-signal-proc, hammond2011graph-wavelets, wang2018lsgcn, huang2021awt-net}, extending them to point cloud sequences is non-trivial due to temporal misalignment and dynamic spatial structures. 
This paper aims to bridge this gap by developing a frequency-aware pipeline for modeling motion patterns in dynamic point clouds.

\section{Methodology} \label{sec:methodology}

As illustrated in Figure~\ref{fig:overview}, our proposed SRENet comprises two core components: the Spectral Decomposition Block (SDeBlock) and the Spectral Re-entry Block (SReBlock).
SDeBlock decomposes an input point cloud sequence into low- and high-frequency components from both temporal and spatial perspectives, and applies self-attention within each frequency band to encode semantics at different resolutions. 
SReBlock takes the obtained spectral features and re-projects them onto a temporal frequency basis, recovering temporal dynamics that have been entangled during cross-axis modeling in SDeBlock.
SRENet is learned with our spectral-aware learning strategy  
that applies spectral contrastive learning separately to the low- and high-frequency spaces with a spectral curriculum schedule to gradually shift the learning focus from low- to high-frequency components,  
mimicking a structural-to-detailed learning process.

\subsection{Problem Formulation}
Following \cite{fan2021p4transformer} and \cite{fan2022pst-transformer}, we first apply the K-nearest neighbor (KNN) algorithm to aggregate local points into patches, and then apply a 4D convolution \cite{fan2021p4transformer} to capture short-term spatio-temporal dependencies between patches in adjacent frames.
We denote the embedded features of a point cloud sequence as $\mathbf{X} \in \mathbb{R}^{T\times P\times C}$,  $\mathbf{y}$ as the ground-truth one-hot label vector, where $T$ is the sequence length, $P$ and $C$ are the number of patches and channels, respectively.
To obtain frame-level observation, a spatial average pooling across patches results in $\bar{\mathbf{X}} \in \mathbb{R}^{T\times C}$. 

\subsection{Spectral Decomposition Block (SDeBlock)} 
\label{section:sdeblock}

For initial spectral modeling, frequency decomposition is performed from both temporal and spatial perspectives with a dual-branch design: {spectral temporal attention} and {spectral spatial attention}, as illustrated in Figure \ref{fig:overview}. 
Each branch follows a unified decomposition-modeling-fusion pipeline, which incorporates wavelet-based decomposition and a self-attention mechanism.
Specifically, the input features are first decomposed via a learnable wavelet transform to extract high- and low-frequency components. Then, a multi-head self-attention mechanism is applied to enable frequency-aware modeling across the decomposed bands. 

\subsubsection{Unified Decomposition-Modeling-Fusion Pipeline}
\label{sec:unified}

\begin{figure}[t]
    \centering
    \includegraphics[width=0.485\textwidth]{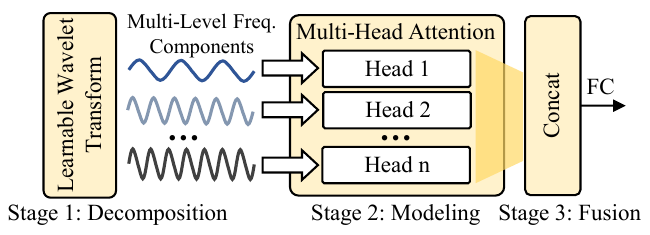}
   \caption{Illustration of the unified Decomposition-Modeling-Fusion pipeline.}
   \label{fig:attention}
\end{figure}

To extract spectral features, the proposed SDeBlock adopts a unified Decomposition-Modeling-Fusion pipeline, serving as the core paradigm integrating temporal and spatial branches. 
\rev{This unified design provides a consistent spectral modeling framework for both the temporal and spatial feature extraction branches.}
The pipeline comprises three stages (Fig. \ref{fig:attention}):

\textbullet Decomposition: To disentangle mixed spectral information in the spatio-temporal space, both branches decompose input features into high-frequency details and low-frequency approximations via a learnable wavelet decomposer, thus facilitating discriminative spectral representation learning.

\textbullet Modeling: Frequency-aware attention is applied to mine inter-frequency dependencies across the above decomposed components.
With these components designated as queries, keys, and values, SDeBlock further enhances frequency-related feature representations.

\rev{\textbullet Fusion: After independent modeling on each frequency band, the resulting features are fused to form comprehensive multi-scale frequency features for subsequent modules.}

\subsubsection{Learnable Wavelet Transform (LWT)}
\label{sec:lwt}

\begin{figure}[t]
    \centering
    \includegraphics[width=0.48\textwidth]{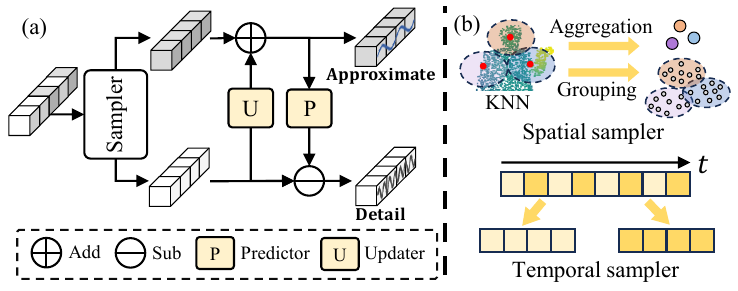}
   \caption{(a) Illustration of the Learnable Wavelet Transform (LWT). (b) The spatial sampler for SSA and the temporal sampler for STA.}
   \label{fig:lwt}
\end{figure}

We propose a learnable wavelet transform (LWT) module (Fig. \ref{fig:lwt}a) for enhanced frequency modeling in dynamic point clouds, simplifying parameter tuning to obtain a suitable wavelet basis\rev{, as point cloud action sequences often contain short-term local motion variations that are difficult to characterize by global spectrum transforms alone. 
Compared with fixed-basis spectral transforms such as the DCT and Fourier transform, and with STFT using a fixed window, wavelet decomposition provides multi-resolution time-frequency localization, which is better suited to capturing transient temporal changes and localized structural variations across multiple time scales.
Moreover, due to the irregular and temporally varying nature of point cloud sequences, we adopt a learnable decomposition in SDeBlock so that the transform basis can adapt to data-dependent dynamics and local structural variations, rather than being constrained by a predefined fixed basis.}

\rev{Specifically, }we adopt a learnable lifting scheme \cite{huang2021awt-net} for both temporal and spatial branches. 
The LWT consists of three key components, as illustrated in Figure \ref{fig:lwt}.

\textbullet Sampler: 
The sampler (Fig. \ref{fig:lwt}b) splits the input signal $\mathbf{X}$ into non-overlapping two subsets $\mathbf{X}_\mathrm{A}$, $ \mathbf{X}_\mathrm{D}$. 
This prevents information loss and provides independent processing channels for subsequent transformation operations.

\textbullet Updater ($U$) and Predictor ($P$):
The updater refines the signal, generating low-frequency approximations;
the predictor reconstructs another input signal and predicts high-frequency details via residual learning. A LWT is formulated as:
\begin{equation}
\begin{split}
\mathbf{\phi}_\text{A} &= \mathbf{X}_\text{A} + U(\mathbf{X}_\text{D}), \\
\mathbf{\phi}_\text{D} &= \mathbf{X}_\text{D} - P(\mathbf{\phi}_\text{A}),
\end{split}
\end{equation}
where $\mathbf{\phi}_\text{A}$,$\mathbf{\phi}_\text{D}$ denote approximation (low-frequency) and detail (high-frequency) 
coefficients, respectively.

For multi-scale frequency analysis, iterative LWT is applied to decompose features into spectral domain components across diverse scales.
Each iteration takes the approximation component of the previous level as input and generates the detail and approximation components of the current level via LWT:
\begin{equation}
\mathbf{\phi}_{\text{A}_l},\mathbf{\phi}_{\text{D}_l} = \text{LWT}(\mathbf{\phi}_{\text{A}_{l-1}}),
\end{equation}
where ${\text{A}_l}$ and ${\text{D}_l}$ denote approximation (low-frequency) and detail (high-frequency) coefficients at level $l$, respectively.
These components provide comprehensive frequency domain information to support subsequent frequency-aware modeling.


\subsubsection{Spectral Temporal Attention (STA)} \label{section:FM-TSB}

The STA module leverages LWT to extract multi-scale temporal frequency components. 
Specifically, LWT of STA adopts even-odd interval temporal sampling to split input features into disjoint subsets, as shown in Figure \ref{fig:lwt}(b);
these subsets are then refined separately by the updater and predictor to yield low- and high-frequency coefficients, respectively.
Given the input $\bar{\mathbf{X}}$, STA applies $L$-level LWT along the temporal dimension, resulting in a set of frequency components:
\begin{equation}
\Phi_\text{STA} = \{ \mathbf{\phi}_\mathrm{D_1}, \mathbf{\phi}_\mathrm{D_2}, \ldots, \mathbf{\phi}_\mathrm{D_L}, \mathbf{\phi}_\mathrm{A_L} \}.
\end{equation}
This can separate dynamic information across different temporal scales, covering short-term local motion mutations to long-term global motion trends.

For frequency-aware temporal modeling, the frequency coefficients serve as the queries $\mathbf{Q}$, keys $\mathbf{K}$ and values $\mathbf{V}$ of a self-attention. 
In detail, for a given decomposition level $\gamma$ and the decomposition coefficient $\bar{\mathbf{X}}_\gamma \in \Phi_\text{STA}$, a self-attention $\text{F-Attention}_\gamma$ is formulated as:
\begin{equation}
\label{equ:concat_f_attention0}
\mathbf{H}_\gamma = \text{F-Attention}_\gamma\bigl(\mathbf{Q}=\bar{\mathbf{X}}_\gamma,\mathbf{K}=\bar{\mathbf{X}}_\gamma,\mathbf{V}=\bar{\mathbf{X}}_\gamma\bigr).
\end{equation}
The outputs from the attention corresponding to each frequency coefficient are concatenated, yielding a set of final outputs $\mathcal{H}=\{ \mathbf{H}_\mathrm{D_1}, \mathbf{H}_\mathrm{D_2},\ldots, \mathbf{H}_\mathrm{D_L}, \mathbf{H}_\mathrm{A_L}\}$.
Finally, these attentions are concatenated to form the final frequency-aware temporal embedding:
\begin{equation}
\label{equ:concat_f_attention}
\mathbf{F}_{\mathrm{STA}} = \mathrm{Concat}( \mathcal{H}) \mathbf{W}_\mathcal{H},
\end{equation}
where $\mathbf{W}_\mathcal{H}$ is a learnable projection matrix.
\subsubsection{Spectral Spatial Attention (SSA)} \label{section:ssa}

To handle the irregular and unordered topology of point clouds, we propose SSA for spatial understanding using a LWT which uses a point cloud-specific spatial sampler. 
Unlike conventional non-learnable transforms, which struggle with the unordered nature of point clouds, this approach enables adaptive spatial decomposition in a permutation-invariant manner.

Specifically, along the spatial dimension of $\mathbf{X}$, the sampler retrieves the K-nearest neighbors $\mathcal{N}_{t,p}$ for each patch $p$ at time step $t$ and applies max-pooling to each patch for feature aggregation, as shown in Figure \ref{fig:lwt}(b). 
A local aggregated feature and a grouped feature can be obtained as:
\begin{equation}
    \mathbf{X}_{t,p}^{\text{group}} = \sum_{t',p'\in \mathcal{N}_{t,p}}{\mathbf{X}_{t',p',:}}, 
    \mathbf{X}_{t,p}^{\text{agg}} = \max_{t',p'\in \mathcal{N}_{t,p}}{\mathbf{X}_{t',p',:}},
\end{equation}
where $\mathbf{X}_{t',p',:}\in\mathbb{R}^{1\times C}$ denotes selecting all channels corresponding to the fixed spatial index $p'$ and the time step $t'$ from $\mathbf{X}$.
To this end, at each time step, the spatial sampler generates initial inputs $\mathbf{X}_{t}^\text{group} = \{\mathbf{X}_{t,p}^\text{group}\}_p$ and $\mathbf{X}_{t}^\text{agg} = \{\mathbf{X}_{t,p}^\text{agg}\}_p$ (both $\mathbb{R}^{P \times C}$) for the LWT.

The approximation branch of LWT leverages aggregated features, as feature aggregation enhances the robustness of smoothed local feature representations. The detail branch adopts grouped features, since group-wise processing effectively preserves high-frequency components in the raw data. The LWT of SSA can be formulated as:
\begin{equation}
\begin{split}
    \mathbf{\psi}_{t,\text{A}} &= \mathbf{X}_{t}^\text{agg} + U(\mathbf{X}_{t}^\text{group}), \mathbf{\psi}_\text{A} = \{\mathbf{\psi}_{t,\text{A}}\}_{t} \in \mathbb{R}^{T \times P \times C}, \\
\mathbf{\psi}_{t,\text{D}} &= \mathbf{X}_{t}^\text{group} - P(\mathbf{\psi}_{t,\text{A}}), \mathbf{\psi}_\text{D} = \{\mathbf{\psi}_{t,\text{D}}\}_{t} \in \mathbb{R}^{T \times P \times C}
\end{split}
\end{equation}
Both the updater ($U$) and predictor ($P$) are designed using fully connected layers (FCs) with a bottleneck architecture, the same as the STA module.

Similar to Eqs.~(\ref{equ:concat_f_attention0})–(\ref{equ:concat_f_attention}) in STA, SSA applies self-attention to both high- and low-frequency components, and concatenates the outputs with a linear projection to form the final frequency-aware spatial embedding $\mathbf{F}_{\mathrm{SSA}}$. 

\subsubsection{STA \& SSA Fusion} \label{section:Fusion}
Finally, we aggregate the outputs from STA and SSA as $\mathbf{F} \in \mathbb{R}^{T \times P \times C}$.
\rev{Specifically, the fusion process is formulated as:
\begin{equation}
\mathbf{F} = \mathrm{FC}\big(\mathrm{Concat}[\mathbf{F}_{\mathrm{STA}} \, \| \, \mathbf{F}_{\mathrm{SSA}}]\big),
\end{equation}
where $\mathrm{Concat}[\cdot \, \| \, \cdot]$ denotes feature-level concatenation along the channel dimension, and $\mathrm{FC}(\cdot)$ denotes a fully connected layer for further modeling.}
Overall, this enables the model to maintain both frequency-aware temporal dynamics and spatial structures.

\subsection{Spectral Re-Entry  (SReBlock)} 
\label{section:re-entry}

After temporal and spatial spectral decomposition and encoding, the fused representation contains nonlinear mixtures of temporal frequency modes.
To re-establish a time-oriented frequency basis for motion understanding, we perform temporal spectral re-projection to comprehensively explore relevant temporal dynamics.
This operation does not repeat the initial decomposition; rather, it re-aligns temporal frequency structures that have been entangled during cross-axis semantic fusion, thereby enabling a learning scheme over well-separated temporal bands and recovering residual temporal patterns critical for action discrimination.

Given fused feature ${\mathbf{F}}$, we first apply spatial max pooling to extract a temporal feature sequence:
\begin{equation}
\bar{\mathbf{F}} = \max_{p} {\mathbf{F}_{:,p,:}} \in \mathbb{R}^{T \times C}.
\end{equation}
This sequence is then decomposed using a one-level Discrete Wavelet Transform (DWT) into low-frequency and high-frequency components:
\begin{equation}
\begin{split}
\mathbf{\omega}_\mathrm{A} &= \mathrm{Pool} \circ \mathrm{Conv1D} \circ \mathrm{DWT}_\mathrm{L}(\bar{\mathbf{F}}), \\
\mathbf{\omega}_\mathrm{D} &= \mathrm{Pool} \circ \mathrm{Conv1D} \circ \mathrm{DWT}_\mathrm{H}(\bar{\mathbf{F}}),
\end{split}
\end{equation}
where $\mathrm{DWT}_\mathrm{L}$ and $\mathrm{DWT}_\mathrm{H}$  denote the low- and high-frequency wavelet basis functions.
$\mathbf{\omega}_\mathrm{A}$ and $\mathbf{\omega}_\mathrm{D} \in \mathbb{R}^{C}$ are compressed low- and high-frequency features, respectively.

\rev{In our implementation, the DWT uses a Daubechies wavelet basis. Compared with the Haar wavelet, higher-order Daubechies wavelets provide improved regularity and more expressive multiscale representations, making them better suited for preserving fine-grained temporal variations. We also note that Symlet wavelets are closely related least-asymmetric variants of the Daubechies family with similar properties. In our preliminary experiments, using Symlet yielded slightly lower recognition accuracy than Daubechies. We therefore adopt the Daubechies basis in the spectral re-entry stage.}

\vspace{-6pt}
\subsection{Spectral-Aware Learning}

The decomposition into low- and high-frequency components enables learning at multiple granularities, allowing spectral-aware learning (SAL) strategies for motion understanding.  
Specifically, our SAL is based on 
(i) a spectral contrastive learning, and  
(ii) a spectral curriculum schedule.

\subsubsection{Spectral Contrastive Learning (SCL)} 
To leverage the distinct semantic characteristics captured by different frequency bands,  
we apply contrastive supervision separately to the high- and low-frequency components.  
\rev{In our framework, SCL is not intended to explicitly model temporal continuity itself. Instead, temporal continuity and dynamic variability are mainly captured by the preceding spectral decomposition and re-entry modules, while SCL serves as an auxiliary supervisory signal on the resulting spectral representations.}

In particular, a contrastive loss~\cite{li2020prototypical} is introduced to enhance class-level discriminability within each frequency subspace.
\rev{We adopt a prototype-based contrastive formulation because our objective here is to enhance class-level discriminability in each spectral subspace, rather than to model temporal evolution directly. Compared with sample-level pairwise contrastive learning, prototype-based supervision is also more stable under noisy point cloud observations and large intra-class variation.}
This loss encourages individual feature $\mathbf{\omega}_\mathrm{A}$ or  $\mathbf{\omega}_\mathrm{D}$ to be close to its corresponding class representation $\bar{\mathbf{\omega}}_\mathrm{A}[\mathbf{y}]$ or $\bar{\mathbf{\omega}}_\mathrm{D}[\mathbf{y}]$  
while being separated from other class prototypes, thereby improving inter-class separability and intra-class compactness.
The contrastive loss is formulated as follows:
\begin{equation}
\begin{aligned}
    \mathcal{L}_{\text{SCL}}(\mathbf{\omega}_\mathrm{A}) &= - \log \frac{\exp(\mathrm{sim}(\mathbf{\omega}_\mathrm{A}, \bar{\mathbf{\omega}}_\mathrm{A}[\mathbf{y}]) / \tau)}
{\sum_{\mathbf{y}'}\limits \exp(\mathrm{sim}(\mathbf{\omega}_\mathrm{A}, \bar{\mathbf{\omega}}_\mathrm{A}[\mathbf{y}']) / \tau)},\\
    \mathcal{L}_{\text{SCL}}(\mathbf{\omega}_\mathrm{D}) &= - \log \frac{\exp(\mathrm{sim}(\mathbf{\omega}_\mathrm{D}, \bar{\mathbf{\omega}}_\mathrm{D}[\mathbf{y}]) / \tau)}
{\sum_{\mathbf{y}'}\limits \exp(\mathrm{sim}(\mathbf{\omega}_\mathrm{D}, \bar{\mathbf{\omega}}_\mathrm{D}[\mathbf{y}']) / \tau)}.
\end{aligned}
\end{equation}
The total loss can be expressed as:
\begin{equation}
\mathcal{L} = \alpha\mathcal{L}_{\text{SCL}}(\mathbf{\omega}_\mathrm{A}) + (1-\alpha)\mathcal{L}_{\text{SCL}}(\mathbf{\omega}_\mathrm{D}) + \beta\mathcal{L}_{\text{CLS}},
\end{equation}
where $\alpha$ and $\beta$ are hyper-parameters and $\mathcal{L}_{\text{CLS}}$ represents a conventional cross-entropy loss for classification taking both low- and high-frequency features.

\subsubsection{Spectral Curriculum Schedule (SCS)} 

Temporal frequency components reflect motion dynamics at varying semantic granularities:  
low-frequency bands typically capture global action trends, while high-frequency bands encode subtle transitions and local variations.  
However, directly optimizing fine-grained objectives in the early stages of training is often unstable and less effective,  
as the model has not yet formed a robust coarse-level understanding.
Inspired by curriculum learning principles, we thus propose a spectral-aware training schedule  
that gradually increases the model's focus on high-frequency components over time.  
Specifically, we control the weight $\alpha$ of the spectral contrastive loss $\mathcal{L}$ in the $t$-th step 
with a linearly decreasing schedule:
\begin{equation}
\alpha_t = \alpha_\text{start} - (\alpha_\text{start} - \alpha_\text{end}) \cdot \frac{t}{\text{\#epoch}},\quad t = 1, 2, ..., \text{\#epoch}.
\end{equation}
This encourages the model to first learn coarse-grained patterns from low-frequency features,  
and then progressively attend to high-frequency residuals,  
resulting in more stable optimization and better overall action discrimination.

\section{Experiments} \label{sec:experiments}
\subsection{Dataset}

\textbf{MSR-Action3D} \cite{li2010msr-action3d} 
focuses on capturing human actions in 3D space, collecting data via depth sensors and covering common daily human actions (e.g., walking, running, jumping, waving hands, clapping, kicking legs, and picking up objects). It includes 20 action classes performed by 10 subjects, totaling 567 videos and 23K frames.

\textbf{NTU-RGBD} \cite{shahroudy2016ntu-rgbd} 
comprises 56,880 real-world videos across 60 categories, performed by 40 subjects and captured from 3 camera views. Evaluation is conducted on two benchmarks: Cross-Subject (CS) and Cross-View (CV), following the same setting in prior works \cite{fan2021p4transformer, fan2022pst-transformer, fan2021pstnet}.
For Cross-Subject, videos from 20 subjects are used for training, while the remaining subjects are reserved for testing. For Cross-View, videos captured from camera views with IDs 2 and 3 are used for training, while those from camera view ID 1 are used for testing. 

\textbf{NTU-RGBD120} \cite{liu2019ntu120}
extends NTU-RGB containing 60 classes, includes 120 action classes and 114K samples. We follow the same protocol as \cite{fan2021p4transformer}, training on 56 subjects and testing on the remaining 50.

\subsection{Implementation Details}

We follow the data preprocessing pipeline in \cite{fan2022pst-transformer} for all datasets. For MSR-Action3D, we sample 2,048 points per frame at an interval of 1, with a spatial radius of 0.3, 64 center points, and 12 KNN neighbors for embedding. For NTU-RGBD, we sample 24 frames per video at a temporal interval of 2, with a spatial radius of 0.1. 

Training is done for 100 epochs on an NVIDIA RTX 2080 Ti using the SGD optimizer, with a linear warmup strategy applied over the first 15 epochs. For MSR-Action3D, we use a batch size of 16 and an initial learning rate of 0.01, decayed by 0.2 at epochs 30, 50, and 70. For NTU-RGBD, the batch size is set to 24, and the decay rate is adjusted to 0.1.

We adopt 5 Transformer layers and 8 attention heads (each 64-dim), with 512 hidden dimensions as in \cite{fan2022pst-transformer}. In 4DConv, the time window is 3 with a step size of 2. 
For STA module, we apply 2-level LWT with 2 heads per component. For SSA module, 1-level LWT is adopted, with both $\mathbf{\psi}_\text{D}$ and $\mathbf{\psi}_\text{A}$ configured with 4 heads and 12 KNN neighbors. For LWT’s updater and predictor, their fully connected hidden dimensions are set to 64. Spectral-Aware Learning uses $\alpha_\text{start}=0.5$ and $\alpha_\text{end}=0.3$.

\begin{table}[ht]
\caption{Recognition Accuracy (\%) on MSR-Action3D. \textit{Trans.} stands for `Transformer'.}
\label{tab:msr-sota}
\begin{center}
\setlength{\tabcolsep}{2mm}{
\begin{tabular}{ccccccc}
\toprule
  \multirow{2}{*}{Method} & \multicolumn{5}{c}{$\#$Frame} & \multirow{2}{*}{Mean} \\
  \cmidrule(lr){2-6}
    & 4 & 8 & 12 & 16 & 24 & \\
\midrule
MeteorNet \cite{liu2019meteornet}    &  78.11 &  81.14 &   86.53 &   88.21 &   88.50 &  84.50 \\
PPTr \cite{wen2022pptr}              &  80.97 &  84.02 &   89.89 &   90.31 &   91.20 &  87.28 \\
PST$^2$ \cite{wei2022pst2}           &  81.14 &  \underline{86.53} &   88.55 &   89.22 &     - &   - \\
PSTNet \cite{fan2021pstnet}          &  81.14 &  83.50 &   87.88 &   89.90 &   91.20 &  86.72 \\
PSTNet++ \cite{fan2021pstnet2}       &  \textbf{81.53} &  83.50 &   88.15 &   90.24 &   92.68 &  87.22  \\
P4Trans. \cite{fan2021p4transformer}     &  80.13 &  83.17 &   87.54 &   89.56 &   90.94 &  86.27\\
{PST-Trans. \cite{fan2022pst-transformer}} &  81.14 &  83.97 &   88.15 &   \underline{91.98} &   \underline{93.73} &  \underline{87.79}\\
Kinet \cite{zhong2022kinet}          &  79.80 &  83.84 &   88.53 &   91.92 &   93.27 &  87.47 \\
Leaf \cite{liu2023leaf} & - & 84.50 & - & 91.50 & 93.84 & - \\
 X4D \cite{jing2024x4d-sceneformer} & - & 86.47 &  - & \textbf{92.56} & 93.90 & - \\
3DInAction \cite{ben20243dinaction}  &  80.47 &  86.20 &   88.22 &   90.57 &   92.23 &  87.54\\
MAMBA4D \cite{liu2024mamba4d}        &  - &    - &     - &     - &   92.68 &   - \\
\midrule
SRENet & \underline{81.48} & \textbf{86.94} & \textbf{89.27} & \underline{91.98} & \textbf{94.07} & \textbf{88.75} \\
\bottomrule
\end{tabular}
}
\end{center}
\end{table}

\begin{table}[t]
\caption{Recognition Accuracy (\%) on NTU-RGBD.
}
\centering
\setlength{\tabcolsep}{2mm}{
\begin{tabular}{cccc}
\toprule
Method & CS & CV & Mean \\
\midrule
3DV-PointNet++ \cite{wang20203dv}                    & 88.8 & 96.3 & 92.6 \\
PSTNet \cite{fan2021pstnet}                          & 90.5 & 96.5 & 93.5 \\
PSTNet++ \cite{fan2021pstnet2}                        & 91.4 & 96.7 & 94.1 \\
P4Transformer \cite{fan2021p4transformer}            & 90.2 & 96.4 & 93.3 \\
PST-Transformer \cite{fan2022pst-transformer}        & 91.0 & 96.4 & 93.7\\ 
Kinet \cite{zhong2022kinet}                         & \underline{92.3} & 96.4 & 94.4\\ 
SequentialPointNet \cite{li2022SequentialPointNet}   & 90.3 & 97.6 & 94.0\\ 
PRENet \cite{he2024prenet}                           & \textbf{93.2} & 97.6 & 95.4 \\ 
KAN-HyperpointNet \cite{chen2024kan-hyperpointNet}  & 91.6 & \textbf{98.4} & 95.0 \\ 
PRG-Net \cite{du2025prg-net}                         & 91.0 & 97.7 & 94.4 \\
\midrule
SRENet & \textbf{93.2} & \underline{98.0} & \textbf{95.6} \\ 
\bottomrule
\end{tabular}
}
\label{tab:ntu-sota}
\end{table}

\begin{table}[t]
\caption{Recognition Accuracy (\%) on NTU-RGBD120. 
}
\centering
\setlength{\tabcolsep}{2mm}{
\begin{tabular}{cc}
\toprule
Method & Accuracy \\
\midrule
3DV-PointNet++ \cite{wang20203dv}                    & 82.4 \\
PSTNet \cite{fan2021pstnet}                          & 87.0 \\
PSTNet++~\cite{fan2021pstnet2}                        & 85.8 \\
P4Transformer \cite{fan2021p4transformer}            & 86.4 \\
PST-Transformer \cite{fan2022pst-transformer}        &\underline{87.5} \\ 
SequentialPointNet \cite{li2022SequentialPointNet}   & 83.5 \\ 
PRG-Net \cite{du2025prg-net}                         & 85.4 \\
\midrule
SRENet & \textbf{87.8} \\ 
\bottomrule
\end{tabular}
}
\label{tab:ntu120-sota}
\end{table}

\subsection{Comparison with State-of-the-Art Methods}
We compare our SRENet with recent state-of-the-art methods. 
The quantitative results in Table \ref{tab:msr-sota} present the 3D action recognition accuracy on MSR-Action3D, evaluated at different frame counts (4, 8, 12, 16, and 24), with the overall mean accuracy reported for each method. 
Our method achieves an average accuracy of 88.75\%, surpassing the previous best method - PST-Transformer \cite{fan2022pst-transformer}, by 0.96\%. 
Compared to previous methods, our SRENet achieves the highest accuracy at 8, 12, and 24 frames, highlighting its effectiveness in both short-term and long-term temporal modeling.
This is due to the spectral decomposition and spectral re-entry mechanism that effectively extracts multi-scale frequency features and enhances the discriminability of temporal representations.

Table \ref{tab:ntu-sota} presents the quantitative results on NTU-RGBD. Our method achieves the best accuracy in both Cross-subject and Cross-view evaluations.
Given that the NTU-RGBD dataset comprises complex actions and scenes requiring fine-grained understanding capabilities, the consistent superiority of our approach across all metrics highlights its effectiveness in handling diverse and challenging scenarios.

Table~\ref{tab:ntu120-sota} shows that our proposed SRENet achieves the best performance on the NTU-RGBD120 dataset under the cross-subject setting.
Our re-implementation of PSTNet++ yields only 85.8\%, indicating its sensitivity to training configurations and potential reliance on specific implementation details. In contrast, our SRENet consistently delivers competitive and stable performance without requiring complex data augmentation or additional structural designs, demonstrating the efficiency and robustness of our Spectral-Aware Learning in modeling the crucial temporal pattern.

\begin{table}[ht]
\caption{
Ablation Study on NTU-RGBD.
}
\centering
\setlength{\tabcolsep}{2.5mm}{
\begin{tabular}{lccc}
\toprule
Method & Params. & CS & CV \\
\midrule
Entire Model &  \rev{29.6M}  & \textbf{93.2} & \textbf{98.0} \\
~- w/o SAL & \rev{29.6M} & 92.5 & 97.5 \\
~- w/o SAL \& SReBlock  & 21.6M & \rev{92.7} & \rev{97.3} \\
~- w/o SAL \& SReBlock \& SDeBlock & 19.9M  & 92.2 & 96.8 \\
\bottomrule
\end{tabular}
}
\label{tab:ablation-total}
\end{table}

\subsection{Ablation Study}
\subsubsection{The Design of SRENet}
We conduct ablation studies to evaluate the contribution of each key component in SRENet, as shown in Table~\ref{tab:ablation-total}.  
Removing the Spectral-Aware Learning (SAL) causes the largest performance drop—especially under the cross-view setting—highlighting its role in learning view-invariant frequency representations.  
Excluding SReBlock also leads to a notable decrease, confirming its importance in re-aligning temporal frequency structures after fusion.  
After removing all core components including the SDeBlock, the model is reduced to a baseline that retains only the standard spatiotemporal attention Transformer backbone, without incorporating any spectral components explicitly. 
Removing SDeBlock results in a noticeable performance degradation (CS: -0.5\%, CV: -0.5\%), highlighting the importance of frequency domain to capture discriminative features.
These results demonstrate that the spectral modules offer complementary benefits, and their joint integration is crucial for the overall effectiveness of SRENet.  
We further analyze the impact of each design choice in detail.

\begin{table}[ht]
\caption{
Ablation Study of SDeBlock on NTU-RGBD.
}
\centering
\setlength{\tabcolsep}{4.3mm}{
\begin{tabular}{lccc}
\toprule
Method & Params. & CS & CV \\
\midrule
SDeBlock Only Model & 21.6M & \textbf{92.7} & \textbf{97.3} \\
~- w/o STA & 20.6M & 92.5 & 97.3 \\
~- w/o SSA & 20.9M & 92.5 & 97.0 \\
~- w/o STA \& SSA & 19.9M & 92.2 & 96.8 \\
\bottomrule
\end{tabular}
}
\label{tab:ablation-sdeblock}
\end{table}

\subsubsection{The Design of SDeBlock}
Table~\ref{tab:ablation-sdeblock} reports an ablation study on SDeBlock to evaluate the roles of its two branches: Spatial Spectral Attention (SSA) and Spectral Temporal Attention (STA).  
To assess the benefit of spectral modeling, we replaced each branch with standard spatial or temporal attention upon removal.  
Disabling either SSA or STA results in a consistent performance drop under both cross-subject and cross-view settings, confirming the individual effectiveness of spatial and temporal frequency modeling.
Notably, removing both branches leads to the largest degradation (CS: -0.5\%, CV: -0.5\%), indicating their complementary nature.  
The full SDeBlock achieves the best results, validating the joint modeling of spatio-temporal spectral patterns.

\begin{table}[ht]
\caption{
Ablation Study of SReBlock and Spectral-Aware Learning (SAL) on NTU-RGBD.
}
\centering
\setlength{\tabcolsep}{4.3mm}{
\begin{tabular}{cc|cc|cc}
\toprule
\multicolumn{2}{c|}{SReBlock} & \multicolumn{2}{c|}{SAL} & \multirow{2}{*}{CS} & \multirow{2}{*}{CV} \\
$\mathbf{\omega}_\mathrm{A}$ & $\mathbf{\omega}_\mathrm{D}$ & SCL & SCS &  &  \\
\midrule
\checkmark & ~ & \checkmark & \checkmark & 92.8 & 97.7 \\
~ & \checkmark & \checkmark & \checkmark & 92.7 & 97.6 \\
\midrule
\checkmark & \checkmark & \checkmark & ~ & 92.9 & 97.8 \\
\checkmark  & \checkmark & \checkmark & \checkmark & \textbf{93.2} & \textbf{98.0} \\
\bottomrule
\end{tabular}
}
\label{tab:ablation-sreblock}
\end{table}

\subsubsection{The Design of SReBlock}
The top two rows of Table~\ref{tab:ablation-sreblock} evaluate the contribution of the SReBlock by selectively enabling low-frequency ($\omega_\mathrm{A}$) and high-frequency ($\omega_\mathrm{D}$) components.  
Disabling either leads to a noticeable performance drop, indicating that both bands provide complementary cues for modeling discriminative temporal dynamics.  
The full version, which integrates both frequency branches, achieves the best performance—demonstrating the effectiveness of dual-band spectral re-alignment in enhancing temporal understanding across frequency scales.

\subsubsection{The Design of Spectral-Aware Learning}
The bottom two rows of Table~\ref{tab:ablation-sreblock} evaluate the contribution of Spectral-Aware Learning.  
Incorporating Spectral Contrastive Learning (SCL) alone yields a clear performance boost over the baseline, demonstrating the value of frequency subspaces optimization for enhancing spectral discriminability.  
This improvement is further amplified when combined with the Spectrum Curriculum Schedule (SCS), which progressively emphasizes different frequency bands and helps guide the model’s optimization path.  
Together, these components contribute to the best overall performance, surpassing the baseline by 1.0\% on CS and 1.2\% on CV. 

\begin{table}[ht]
\caption{
\rev{Comparison of Frequency Decomposition Methods on NTU-RGBD (CS).}
}
\centering
\setlength{\tabcolsep}{0.8mm}{
\rev{
\begin{tabular}{c|ccccc}
\toprule
\multicolumn{6}{c}{Different frequency decompositions in SDeBlock} \\
\midrule
~ & DCT & FFT & DWT & LWT & STFT\\
Accuracy & 92.3 & 92.7 & 93.0 & \textbf{93.2} & 92.9 \\
\midrule
\multicolumn{6}{c}{Different wavelet bases in SReBlock} \\
\midrule
~ & Haar \cite{wave-base-haar1910} & Symlet \cite{wave-base-sym1992} & Daubechies \cite{wave-base-db1988} & Learnable & ~~~~~~~~ \\
Accuracy & 92.6 & 92.9 & \textbf{93.2} &  92.8 & ~ \\
\bottomrule
\end{tabular}
}
}
\label{tab:ablation-wave-comparison}
\end{table}

\subsubsection{\rev{Validation of Frequency Decomposition Strategies}}
\rev{
Table~\ref{tab:ablation-wave-comparison} empirically validates the frequency decomposition choices introduced in SDeBlock and SReBlock. In the upper part, wavelet-based decompositions perform better overall, and LWT achieves the best performance. This suggests that SDeBlock benefits from learnable decomposition, since it operates on earlier features with more irregular and data-dependent temporal variations, for which adaptive bases are more suitable.
}

\rev{
The lower part of Table~\ref{tab:ablation-wave-comparison} further validates the wavelet-basis choice in SReBlock.
Among the tested options, Daubechies achieves the best performance, while the learnable basis does not provide further improvement. 
This indicates that the spectral re-entry stage relies more on stable spectral re-alignment, making the fixed Daubechies basis better suited for recovering a clearer frequency structure, whereas a learnable decomposition may weaken this effect by introducing additional instability into the re-alignment process.
}

\begin{table}[ht]
\caption{Studies of Efficiency on NTU-RGBD (CS).}
\centering
\setlength{\tabcolsep}{3.9mm}{
\begin{tabular}{lccc}
\toprule
Method & Params. & Time (ms) & Accuracy \\
\midrule
P4Transformer      & 44.2M & 264 & 90.2 \\
PST-Transformer   & 44.2M & 266 & 91.0 \\
SRENet (ours) & 29.6M & 256 & \textbf{93.2} \\
~-w/o re-entry & 21.6M & 253 & 92.8 \\
\bottomrule
\end{tabular}
}
\label{tab:efficiency}
\end{table}

\begin{table}[ht]
\caption{\rev{
Inference Time (ms) under Different Input Sequence Lengths on NTU-RGBD.
}}
\centering
\scalebox{1.0}{
\setlength{\tabcolsep}{3.518mm}{
\rev{
\begin{tabular}{c|ccccc}
\toprule
\#Frame & 12 & 24 & 32 & 48 & 64 \\
\midrule
PST-Transformer & 150 & 266 & 436 & 857 & 1340 \\
SRENet (ours)   & 149 & 253 & 331 & 481 & 625 \\
\bottomrule
\end{tabular}
}
}
}
\label{tab:efficiency-len}
\end{table}

\subsubsection{Efficiency}
Table~\ref{tab:efficiency} compares the computational efficiency of our SRENet with previous Transformer-based methods, evaluated on a single NVIDIA RTX 2080 Ti GPU.  
We maintain the same total dimension for the spatial and temporal branches as in previous methods.
Thanks to a smaller hidden layer dimension, our SRENet uses fewer parameters (29.6M) compared to existing Transformer-based methods while achieving the highest accuracy with a processing time of 256 ms, demonstrating both efficiency and high performance.
For $T\times N$ tokens, previous methods have an attention computation complexity of $O(T^2N^2)$, whereas our spatio-temporal separable structure reduces it to $O(T^2N+TN^2)$. 
\rev{This advantage becomes more evident for longer input sequences, as shown in Table~\ref{tab:efficiency-len}. 
As the sequence length increases, SRENet exhibits a substantially slower increase in inference time than PST-Transformer, demonstrating better scalability for longer input sequences.
}

\begin{table}[t]
\caption{\rev{Sensitivity Analysis of LWT on MSR-Action3D.}}
\centering
\scalebox{1.0}{
\setlength{\tabcolsep}{2.95mm}{
\rev{
\begin{tabular}{c|ccc}
\toprule
Decomposition level of LWT (STA) & 1 & 2 & 3 \\
\midrule
Accuracy & 88.89 & \textbf{89.27} & 87.88 \\
\bottomrule
\end{tabular}
}
}
}

\vspace{2pt}

\scalebox{0.92}{
\setlength{\tabcolsep}{0.98mm}{
\rev{
\begin{tabular}{c|cccccc}
\toprule
Neighborhood size of LWT (SSA) & 2 & 4 & 8 & 12 & 16 & 24\\
\midrule
Accuracy & 87.88 & 88.89 & 89.23 & \textbf{89.27} & \textbf{89.27} & 89.23 \\
\bottomrule
\end{tabular}
}
}
}

\vspace{2pt}

\scalebox{1.0}{
\setlength{\tabcolsep}{2.1mm}{
\rev{
\begin{tabular}{c|ccccc}
\toprule
Bottleneck dim of LWT & 16 & 32 & 64 & 128 & 256 \\
\midrule
Accuracy & 88.55 & 89.23 & 89.27 & 89.15 & \textbf{89.60} \\
\bottomrule
\end{tabular}
}
}
}
\label{tab:ablation-lwt}
\end{table}

\subsection{\rev{Hyperparameter Setting}}
\subsubsection{\rev{Hyperparameter Analysis of LWT}}
\rev{
Table~\ref{tab:ablation-lwt} presents the hyperparameter analysis of LWT under the 12-frame setting.
The top rows compare STA with different LWT decomposition levels. Due to downsampling, the level of LWT is limited by the input feature maps' size. The STA’s LWT levels are limited to 1-3, while that for the SSA was set to only 1. LWT with 2 levels achieved the highest performance, indicating that increasing the LWT levels enables more effective extraction of multi-scale frequency features. 3-level LWT leads to performance degradation, which could stem from high decomposition complexity that causes feature redundancy and undermines the completeness of action features.
}

\rev{
The middle and bottom rows of Table~\ref{tab:ablation-lwt} show the effects of the KNN neighborhood size in SSA and the bottleneck dimension in LWT, respectively. 
As the neighborhood size increases, the performance generally improves, suggesting that a larger local region helps capture richer spatial context. Small bottleneck dimensions already work well because the bottleneck helps suppress noise.
Considering both performance and efficiency, we set them to 12 and 64, respectively.
}

\begin{table}[ht]
\caption{
Hyperparameter Analysis of Spectral Curriculum Schedule (SCS) factor $\alpha$ on NTU-RGBD (CS).
}
\centering
\scalebox{1.0}{
\setlength{\tabcolsep}{2.45mm}{
\begin{tabular}{c|ccccccc}
\toprule
$\alpha$    & 0.0   & 0.2   & 0.4   & 0.5    & 0.6  & 0.8    & 1.0 \\
\midrule
Accuracy    & 92.7  & 92.6  & 92.7  &\textbf{92.9}   &92.8   &\textbf{92.9} & 92.8 \\
\bottomrule
\end{tabular}
}
}
\vspace{2pt}

\scalebox{0.935}{
\setlength{\tabcolsep}{0.985mm}{
\begin{tabular}{c|c|ccc|ccc}
\toprule
($\alpha_\text{start}$, $\alpha_\text{end}$)
& (0.5, 0.5) & (0.5,0.3) & (0.3,0) & (1,0.5) &
(0.5,1) & (0.3,0.5) & (0,0.5) \\
\midrule
Accuracy & 
92.9 & \textbf{93.2} & 92.9 & 92.6 &
93.0 & 92.9 & 92.9 \\
\bottomrule
\end{tabular}
}
}
\label{tab:ablation-params-sal-factor}
\end{table}

\subsubsection{Hyperparameter Analysis of SCS}
The top row of Table \ref{tab:ablation-params-sal-factor} demonstrates the contribution of hyperparameter $\alpha$ in Spectral Curriculum Schedule (SCS).
Performance improves with increasing $\alpha$, peaking at $\alpha \ge 0.5$.
This indicates that enhancing low-frequency weighting helps SRENet extract key action recognition information.

The bottom row of Table \ref{tab:ablation-params-sal-factor} compares the impact of different combinations ($\alpha_\text{start}$, $\alpha_\text{end}$). 
Optimal performance is achieved as $\alpha$ drops from 0.5 to 0.3, which verifies that prioritizing low-frequency feature learning followed by high-frequency features enhances performance. However, extreme or mismatched $\alpha_\text{start}$ and $\alpha_\text{end}$ values lead to less ideal accuracy.

\begin{table}[t]
\caption{
\rev{Hyperparameter Analysis of the Temperature $\tau$ in Spectral Contrastive Learning (SCL) on NTU-RGBD (CS).}
}
\centering
\scalebox{1.0}{
\setlength{\tabcolsep}{3.12mm}{
\rev{
\begin{tabular}{c|cccccc}
\toprule
$\tau$ & 0.1 & 0.25 & 0.5 & 1.0 & 2.0 & 4.0 \\
\midrule
Accuracy & 92.3 & 92.1 & 92.5 & \textbf{93.2} & 92.7 & 92.8 \\
\bottomrule
\end{tabular}
}
}
}
\label{tab:ablation-param-temp}
\end{table}

\begin{table}[t]
\caption{
\rev{Hyperparameter Analysis of the Number of Attention Heads on MSR-Action3D.}
}
\centering
\setlength{\tabcolsep}{2.43mm}{
\rev{
\begin{tabular}{c|cccccc}
\toprule
Head & 2 & 4 & 6 & 8 & 12 & 16\\
\midrule
Accuracy & 86.20 & 88.89 & 89.23 & 89.27 & 89.60 & \textbf{89.87} \\
\bottomrule
\end{tabular}
}
}
\label{tab:ablation-param-atten-head}
\end{table}

\subsubsection{\rev{Temperature of SCL}}
\rev{
Table~\ref{tab:ablation-param-temp} presents the effect of the temperature parameter $\tau$ in SCL. 
The best recognition performance is achieved at $\tau = 1.0$, indicating that a moderate temperature provides a better balance between optimization stability and spectral discriminability.
}

\subsubsection{\rev{Attention Head}}
\rev{
Table~\ref{tab:ablation-param-atten-head} compares different numbers of attention heads under the 12-frame setting. 
Increasing the number of attention heads generally improves accuracy but also increases model cost. Based on this trade-off, we report the 8-head model's performance compared with other models.
}

\begin{table}[ht]
\caption{
Recognition Accuracy (\%) on NTU-RGBD actions of different difficulty levels.
}
\centering
\setlength{\tabcolsep}{4.8mm}{
\begin{tabular}{ccccc}
\toprule
Method & Hard & Medium & Easy\\
\midrule
P4Transformer \cite{fan2021p4transformer}      & 71.3 & 85.7 & 95.1 \\
PST-Transformer \cite{fan2022pst-transformer}   & 71.8 & 86.1 & 95.4 \\
PSTNet \cite{fan2021pstnet} & \underline{72.4} & \underline{86.4} & \underline{95.5} \\
SRENet (ours) & \textbf{80.2} & \textbf{90.0} & \textbf{96.9} \\
\midrule
$\Delta$  & 7.8 & 3.6 & 1.4 \\
\bottomrule
\end{tabular}
}
\label{tab:fine-grain-set}
\end{table}

\begin{figure}[ht]
    \centering
    \includegraphics[width=0.45\textwidth]{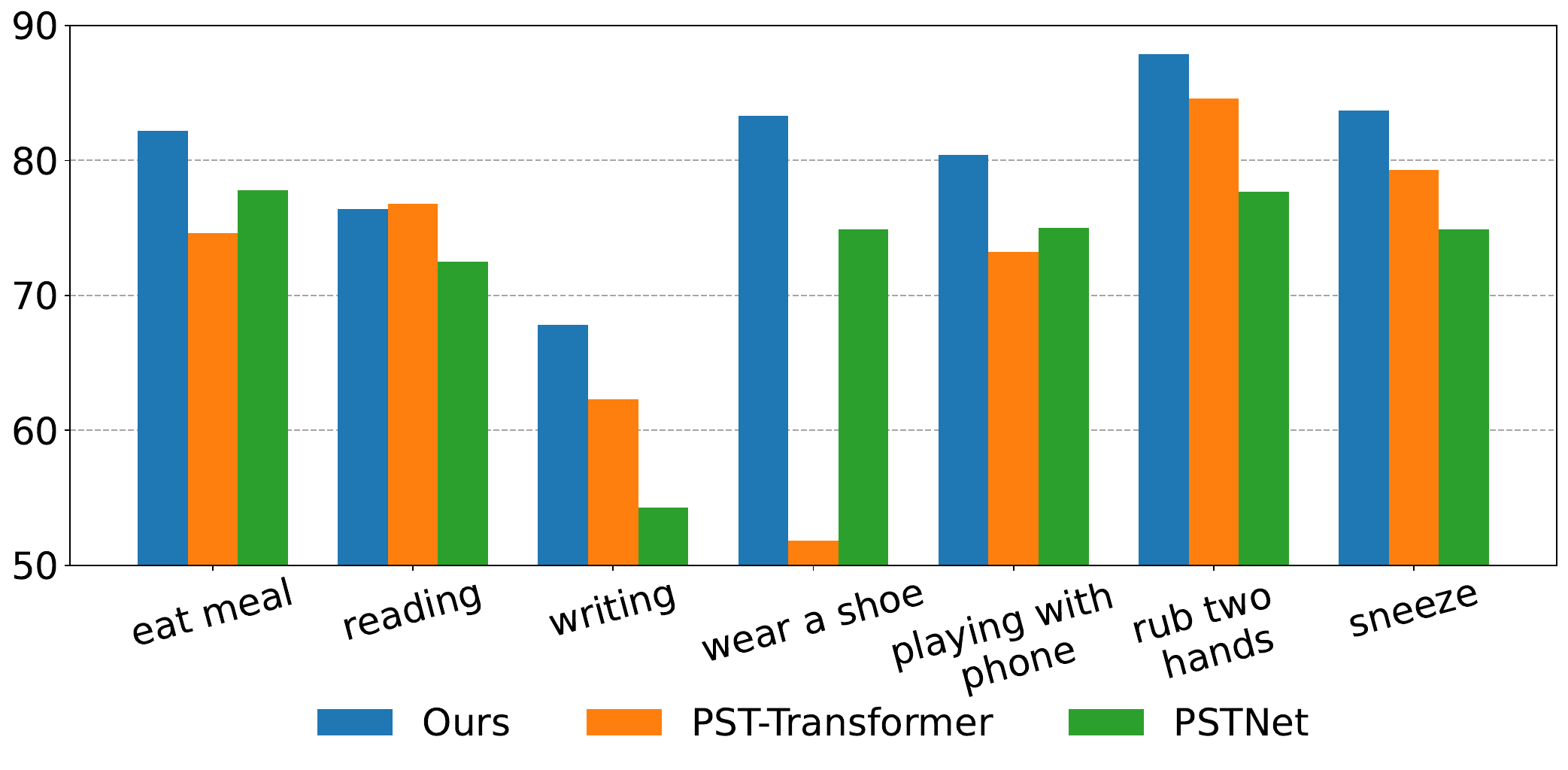}
    \caption{Recognition accuracy (\%) comparison in hard set.}
   \label{fig:hard-set}
\end{figure}

\vspace{-4mm}

\subsection{Performance on fine-grained Actions}
We evaluate our SRENet in distinguishing fine-grained actions with subtle differences. 
Following the settings of \cite{wu2024freqmixformer}, we partition NTU-RGBD actions into three sets according to the accuracy of PSTNet \cite{fan2021pstnet}: actions with accuracy below 80\% are assigned to the hard set, those between 80\% and 90\% to the medium set, and those above 90\% to the easy set. All fine-grained actions—such as \textit{writing}, \textit{reading}, and \textit{eating meal}—are categorized into the hard set.

We compare our SRENet with PSTNet, P4Transformer and PST-transformer, as shown in Table \ref{tab:fine-grain-set}. SRENet achieves the optimal performance across all difficulty levels, with a remarkable improvement on the hard set.
Figure \ref{fig:hard-set} visualizes the hard-set accuracy comparisons — for most action categories, our SRENet achieves the highest accuracy, especially in action \textit{writing} and \textit{wear a shoe} (action differences are subtler), where the accuracy gap is more obvious. 
These results demonstrate our SRENet’s strength in capturing fine-grained features for robust performance in challenging scenarios.

\begin{table}[ht]
\caption{
\rev{
Few-shot Recognition Accuracy (\%) on NTU-RGBD120 with 95\% Confidence Interval.
}
}
\centering
\setlength{\tabcolsep}{1.65mm}{
\rev{
\begin{tabular}{lccc}
\toprule
Method & 1-shot & 4-shot & 8-shot \\
\midrule
PST-transformer \cite{fan2022pst-transformer} & 54.55 $\pm$ 3.87 & 70.58 $\pm$ 2.12 & 76.68 $\pm$ 0.69 \\
SRENet (ours)         & \textbf{68.53 $\pm$ 3.82} & \textbf{82.76 $\pm$ 0.38} & \textbf{87.15 $\pm$ 0.31} \\
- w/o re-entry  & 66.62 $\pm$ 2.74 & 80.80 $\pm$ 0.47 &  85.06 $\pm$ 0.31\\
\bottomrule
\end{tabular}
}
}
\label{tab:sota-few-shot}
\end{table}

\vspace{-2mm}

\subsection{\rev{Performance on Few-shot Recognition}}
\rev{
Few-shot action recognition has been extensively studied in conventional video-based action understanding \cite{qu2026few-shot-llm-st, qu2025few-shot-mvp}. 
Likewise, we further examine the effectiveness of our SRENet under limited supervision for point cloud action recognition.
We evaluate SRENet under 1/4/8-shot settings on NTU-RGBD120. The evaluation protocol is as follows: 100 action classes are used for base training, while the remaining 20 unseen classes are used for few-shot evaluation. For each unseen class, 1, 4, or 8 labeled samples are selected as the support set, and recognition is performed by dot-product similarity matching on the final output representations.
The results in Table \ref{tab:sota-few-shot} show that SRENet consistently outperforms the baseline across all few-shot settings, demonstrating its effectiveness under limited supervision. We attribute this improvement to the more transferable spectral representation learned by our method, which disentangles global trends and subtle local dynamics and further re-aligns temporal frequency structure after fusion.
}

\begin{figure*}[t]
    \centering
    \includegraphics[width=0.985\textwidth]{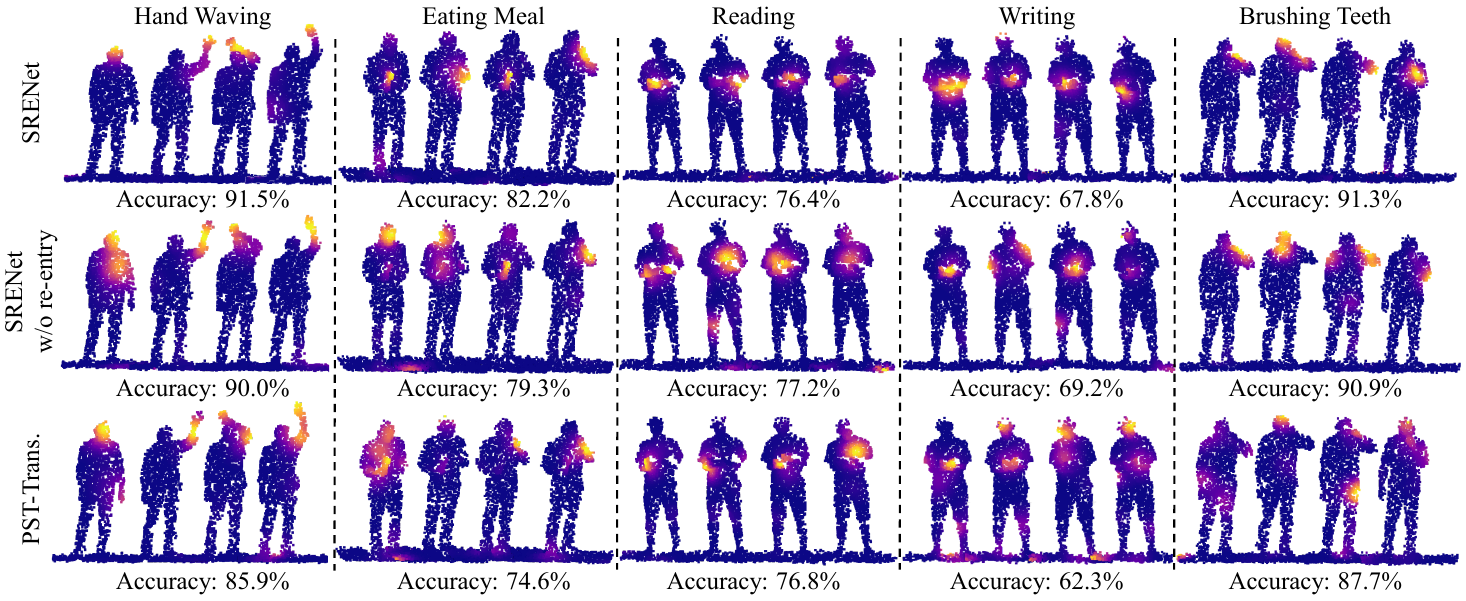} 
    \caption{
    Comparison of class activation maps \rev{
        on NTU-RGBD with class-wise accuracies.
    }
}
   \label{fig:vis-cam-cmp}
\end{figure*}

\subsection{Visualization}
\subsubsection{Activation Maps}
\begin{figure}[t]
    \centering
    \includegraphics[width=0.46\textwidth]{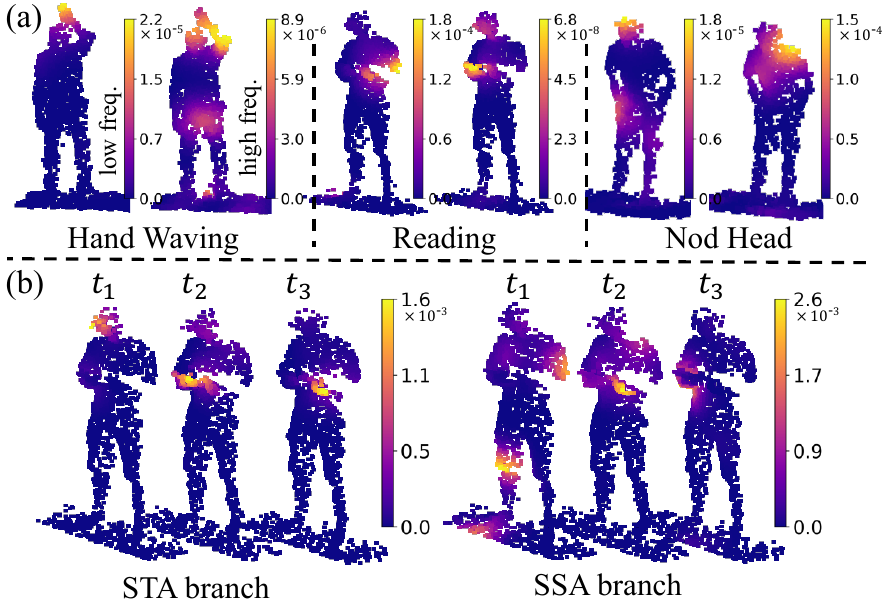} 
    \caption{
    \rev{
        Class Activation Map (CAM).
    (a) CAMs of low- (left) and high-frequency (right) components in the final attention layer.
    (b) CAMs of STA and SSA branches for the action “\textit{reading}”.
    Brighter areas indicate higher activation scores.
    }
}
   \label{fig:vis-cam}
\end{figure}

Figure~\ref{fig:vis-cam} presents class activation maps generated using GradCAM~\cite{selvaraju2017gradcam}.  
As shown in Figure~\ref{fig:vis-cam}(a), low-frequency features tend to activate on body regions exhibiting slower, larger-scale motion—such as the arms during \textit{hand waving}, the hands during \textit{reading}, and the head nodding motion in \textit{nod head}—highlighting their capacity to capture global motion trends.  
In contrast, high-frequency features emerge in regions associated with subtle or rapid movements, such as the hips in \textit{hand waving}, the facial region in \textit{reading}, and the torso in \textit{nod head}.  
These observations suggest that low-frequency components encode coarse, structural motion patterns, while high-frequency activations reflect localized, fine-grained dynamics often arising in non-rigid or auxiliary body parts.

These observations align with the activation patterns in the figure, supporting the effectiveness of spectral modeling.
Figure~\ref{fig:vis-cam}(b) illustrates the distinct modeling behaviors of the STA and SSA branches.
The STA branch generates temporally consistent activations, indicating that it captures motion patterns evolving over time—such as sustained gestures or sequential pose transitions. 
In contrast, the SSA branch responds sharply to frame-level changes, as it focuses on local geometric configurations in individual frames. The resulting spatial saliency shifts more abruptly, revealing its sensitivity to pose-specific spatial features.
This divergence reflects the architectural complementarity.

Figure \ref{fig:vis-cam-cmp} presents the CAMs of our SRENet and PST-Transformer.
As shown in the top row (SRENet), activated regions consistently localize to task-relevant body parts: the swinging arm in \textit{hand waving}, the utensil-holding hand in \textit{eating}, and the pen-holding hand in \textit{writing}. Removing the reentry modules results in dispersed (e.g., spreading to the torso in \textit{hand waving}) or misaligned (e.g., weakened hand signals in \textit{writing}) activations, demonstrating that this module enhances task-specific saliency.
\rev{
The annotated accuracies in the figure further support this observation, with the full model achieving better or comparable results on most actions.
}

In contrast, PST-Transformer (bottom row) shows weaker activation focus: it activates both irrelevant shoulder regions and the brushing hand in tooth brushing, and diffuses attention across the entire upper body instead of the book-holding hand in reading. 
These differences confirm that our SRENet better prioritizes fine-grained, task-critical spatiotemporal features.

\begin{figure}[t]
    \centering
    \includegraphics[width=0.46\textwidth]{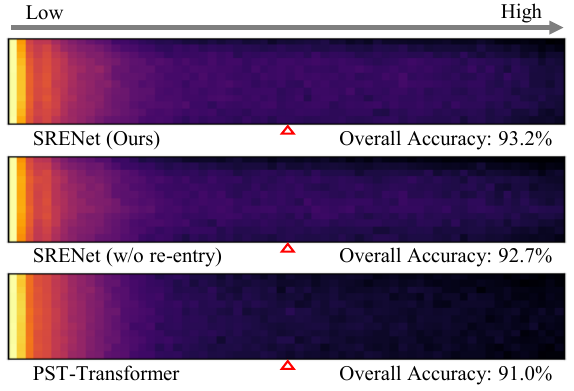} 
    \caption{The frequency amplitude map (64$\times$12) of the final feature map of PST-Transformer and our method, \rev{with overall accuracies on NTU-RGBD (CS)}. The horizontal represents frequency from low to high, with brighter colors indicating higher amplitudes. 
    High-frequency energy ratio (right of red triangle): SRENet (18.6\%), SRENet w/o re-entry (16.0\%) and PST-Transformer (15.6\%).
    }
   \label{fig:vis-freq}
\end{figure}

\subsubsection{Spectrum Analysis}
Figure \ref{fig:vis-freq} shows the amplitude map in the frequency domain.
Due to the unordered nature of points in the sequence, we adopt the Graph Spectral Transform \cite{shuman2013graph-signal-proc} 
to convert the output feature map into the frequency domain. The amplitude map is obtained by randomly selecting 100 samples and averaging the features from 32 randomly selected channels. 
We compare our SRENet with the state-of-the-art Transformer-based PST-Transformer~\cite{fan2022pst-transformer} under a similar architecture but without frequency modeling to highlight the differences.
The superior representation of high-frequency signals in our method—compared to PST-Transformer and SRENet without SReBlock—is attributed to the frequency-aware modeling, which explicitly preserves spectral details. 

\subsubsection{Representation Visualization}
\begin{figure}[t]
  \centering
    \subfloat[SRENet]{
    \includegraphics[width=0.23\textwidth]{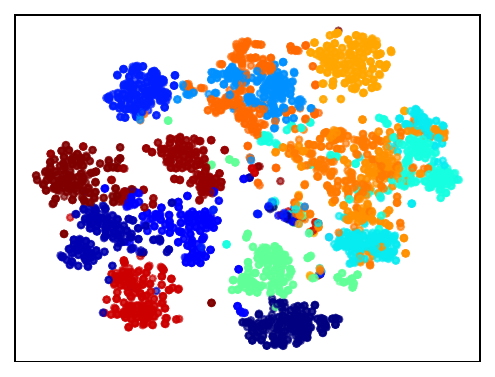}
    }
    \subfloat[PST-Transformer]{
    \includegraphics[width=0.23\textwidth]{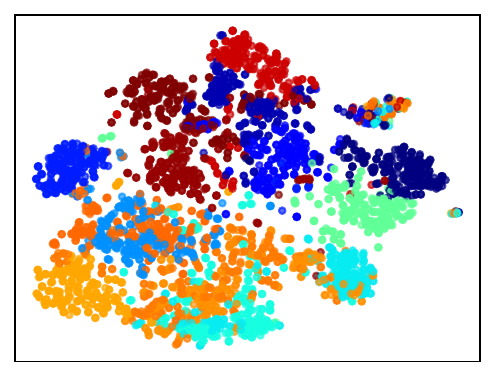}
    }
    \caption{The t-SNE visualization on NTU-RGBD.}
    \label{fig:tsne}
\end{figure}

We further visualize the feature distribution using t-SNE, as shown in Figure \ref{fig:tsne}. 
To evaluate feature effectiveness, we select 15 NTU-RGBD categories that exhibit fine-grained patterns and pose challenges for discriminative representation. 
Compared to the PST-Transformer, which equips with conventional attention mechanisms, our SRENet exhibits more distinct class boundaries in the feature space, indicating its superior discriminability. This advantage is attributed to our spectral-aware design, which extracts a richer set of spectral components to capture subtle changes in human motion.

\subsection{\rev{Limitations and Future Work}}
\subsubsection{\rev{Limitations}}
\begin{figure}[t]
    \centering
    \includegraphics[width=0.45\textwidth]{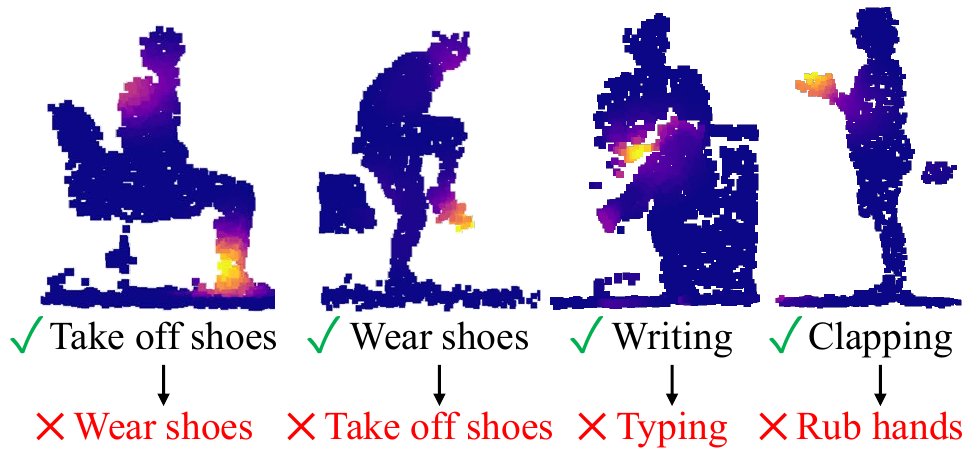} 
    \caption{\rev{
    Some failure cases with CAM visualizations. These examples are selected from the hard and medium sets. \textcolor{green!75!black}{$\checkmark$} denotes the ground-truth labels, and \textcolor{red}{$\times$} denotes the incorrect predictions.
    }
    }
   \label{fig:limit}
\end{figure}

\rev{
Figure \ref{fig:limit} presents several representative misclassified examples and their CAM visualizations. The results show that the proposed method can still struggle with fine-grained action discrimination when different classes share highly similar global motion patterns and only differ in subtle local dynamics, such as local motion direction (\textit{wear shoes} versus \textit{take off shoes}), hand-centered movement details (\textit{typing} versus \textit{writing}), or contact patterns (\textit{clapping} versus \textit{rub hands}).
}
\subsubsection{\rev{Future Work}}
\rev{
To address the limitations, in future work, we will further explore:
1) incorporating more explicit temporal-order and motion-direction modeling to better distinguish actions that share similar global motion patterns but differ in execution order or directional evolution; 
2) enhancing fine-grained local motion modeling, particularly for hand-centered dynamics, to capture subtle movement differences that are easily obscured in global representations; 
3) introducing more explicit modeling of contact and interaction patterns, such as hand-hand or hand-object relations, to improve discrimination between actions with similar coarse trajectories but different local interaction semantics.
And 4) zero-shot recognition strategies for unseen action categories.
}

\rev{
Moreover, although SRENet is developed for point cloud action recognition, its significance is not limited to this task. To the best of our knowledge, this work is the first attempt to introduce frequency modeling into sequential point cloud analysis, providing a new perspective beyond conventional spatio-temporal modeling. With this framework, it may inspire future research in relevant applications, such as \revtwo{gait recognition \cite{zheng2022gait-reg-wild},} group activity recognition \cite{yan2020group-action-recg}, compositional action recognition \cite{yan2023compsi-action-recg}, and event localization \cite{chen2026av-ssan, xing2025locality}. 
}

\section{Conclusion} \label{sec:conclusion}

We present SRENet, a novel spectral-aware framework for action recognition in point cloud sequences. 
By incorporating SDeBlock and SReBlock, SRENet effectively disentangles low- and high-frequency components for frequency-specific modeling.  
The spectral-aware learning strategy further enhances spectral discrimination via low- and high-frequency subspaces.
Extensive experiments demonstrate SRENet's state-of-the-art performance.

\bibliographystyle{IEEEtran}
\bibliography{IEEEabrv,main}

@String(CVPR= {IEEE Conf. Comput. Vis. Pattern Recog.})

@String(ICCV= {Int. Conf. Comput. Vis.})

@String(ICASSP=	{ICASSP})

@String(AAAI = {AAAI})

@String(CVPR  = {CVPR})

@String(ICCV  = {ICCV})

@inproceedings{fan2021pstnet,
    title={P{S}{T}Net: Point Spatio-Temporal Convolution on Point Cloud Sequences},
    author={Hehe Fan and Xin Yu and Yuhang Ding and Yi Yang and Mohan Kankanhalli},
    booktitle={Proc. Int. Conf. Learn. Represent.},
    pages={18296--18318},
    year={2021}
}

@inproceedings{wang20203dv,
  title={3{D}{V}: 3{D} dynamic voxel for action recognition in depth video},
  author={Wang, Yancheng and Xiao, Yang and Xiong, Fu and Jiang, Wenxiang and Cao, Zhiguo and Zhou, Joey Tianyi and Yuan, Junsong},
  booktitle={Proc. IEEE/CVF Conf. Comput. Vis. Pattern Recognit. (CVPR)},
  pages={511--520},
  year={2020}
}

@inproceedings{fan2021p4transformer,
  title={Point 4{D} transformer networks for spatio-temporal modeling in point cloud videos},
  author={Fan, Hehe and Yang, Yi and Kankanhalli, Mohan},
  booktitle={Proc. IEEE/CVF Conf. Comput. Vis. Pattern Recognit. (CVPR)},
  pages={14204--14213},
  year={2021}
}

@inproceedings{liu2019meteornet,
  title={Meteornet: Deep learning on dynamic 3{D} point cloud sequences},
  author={Liu, Xingyu and Yan, Mengyuan and Bohg, Jeannette},
  booktitle={Proc. IEEE/CVF Int. Conf. Comput. Vis.},
  pages={9246--9255},
  year={2019}
}

@article{fan2022pst-transformer,
  title={Point spatio-temporal transformer networks for point cloud video modeling},
  author={\vspace{0mm}Fan, Hehe and Yang, Yi and \vspace{0mm}Kankanhalli, Mohan},
  journal={IEEE Trans. Pattern Anal. Mach. Intell.},  
  volume={45},
  number={2},
  pages={2181--2192},
  year={2022},
  publisher={IEEE}
}

@article{fan2021pstnet2,
  title={Deep hierarchical representation of point cloud videos via spatio-temporal decomposition},
  author={Fan, Hehe and Yu, Xin and Yang, Yi and Kankanhalli, Mohan},
  journal={IEEE Trans. Pattern Anal. Mach. Intell.},
  volume={44},
  number={12},
  pages={9918--9930},
  year={2021},
  publisher={IEEE}
}

@inproceedings{liu2024mamba4d,
  title={MAMBA4{D}: Efficient 4{D} Point Cloud Video Understanding with Disentangled Spatial-Temporal State Space Models},
  author={Liu, Jiuming and Han, Jinru and Liu, Lihao and Aviles-Rivero, Angelica I and Jiang, Chaokang and Liu, Zhe and Wang, Hesheng},
  booktitle={Proc. IEEE/CVF Conf. Comput. Vis. Pattern Recognit. (CVPR)},
  pages={17626--17636},
  year={2025}
}

@inproceedings{wen2022pptr,
  title={Point primitive transformer for long-term 4{D} point cloud video understanding},
  author={Wen, Hao and Liu, Yunze and Huang, Jingwei and Duan, Bo and Yi, Li},
  booktitle={Proc. Eur. Conf. Comput. Vis.},
  pages={19--35},
  year={2022},
  organization={Springer}
}

@inproceedings{zhong2022kinet,
  title={No pain, big gain: classify dynamic point cloud sequences with static models by fitting feature-level space-time surfaces},
  author={Zhong, Jia-Xing and Zhou, Kaichen and Hu, Qingyong and Wang, Bing and Trigoni, Niki and Markham, Andrew},
  booktitle={Proc. IEEE/CVF Conf. Comput. Vis. Pattern Recognit. (CVPR)},
  pages={8510--8520},
  year={2022}
}

@inproceedings{ben20243dinaction,
  title={3{D}in{A}ction: Understanding human actions in 3{D} point clouds},
  author={Ben-Shabat, Yizhak and Shrout, Oren and Gould, Stephen},
  booktitle={Proc. IEEE/CVF Conf. Comput. Vis. Pattern Recognit. (CVPR)},
  pages={19978--19987},
  year={2024}
}

@inproceedings{wu2024freqmixformer,
  title={Frequency Guidance Matters: Skeletal Action Recognition by Frequency-Aware Mixed Transformer},
  author={Wu, Wenhan and Zheng, Ce and Yang, Zihao and Chen, Chen and Das, Srijan and Lu, Aidong},
  booktitle={Proc. 32nd ACM Int. Conf. Multimedia},  
  pages={4660--4669},
  year={2024}
}

@inproceedings{gao2023freq-motion,
  title={Decompose more and aggregate better: Two closer looks at frequency representation learning for human motion prediction},
  author={Gao, Xuehao and Du, Shaoyi and Wu, Yang and Yang, Yang},
  booktitle={Proc. IEEE/CVF Conf. Comput. Vis. Pattern Recognit. (CVPR)},
  pages={6451--6460},
  year={2023}
}

@inproceedings{pan2022hilo-vit,
  title={Fast vision transformers with hilo attention},
  author={Pan, Zizheng and Cai, Jianfei and Zhuang, Bohan},
  booktitle={Proc. Adv. Neural Inf. Process. Syst.},
  volume={35},
  pages={14541--14554},
  year={2022}
}

@inproceedings{si2022inception-transformer,
  title={Inception transformer},
  author={Si, Chenyang and Yu, Weihao and Zhou, Pan and Zhou, Yichen and Wang, Xinchao and Yan, Shuicheng},
  booktitle={Proc. Adv. Neural Inf. Process. Syst.},
  volume={35},
  pages={23495--23509},
  year={2022}
}

@inproceedings{park2022how-vit-work,
  title={How Do Vision Transformers Work?},
  author={Namuk Park and Songkuk Kim},
  booktitle={Proc. Int. Conf. Learn. Represent.},
  pages={4287--4312},
  year={2022}
}

@inproceedings{wei2022pst2,
  title={Spatial-temporal transformer for 3{D} point cloud sequences},
  author={Wei, Yimin and Liu, Hao and Xie, Tingting and Ke, Qiuhong and Guo, Yulan},
  booktitle={Proc. IEEE/CVF Winter Conf. Appl. Comput. Vis. (WACV)},  
  pages={1171--1180},
  year={2022}
}

@inproceedings{chen2022maple,
  title={M{A}{P}{L}{E}: Masked pseudo-labeling autoencoder for semi-supervised point cloud action recognition},
  author={Chen, Xiaodong and Liu, Wu and Liu, Xinchen and Zhang, Yongdong and Han, Jungong and Mei, Tao},
  booktitle={Proc. ACM Int. Conf. on Multimedia},
  pages={708--718},
  year={2022}
}

@inproceedings{vaswani2017transformer,
  title={Attention is all you need},
  author={Vaswani, A},
  booktitle={Proc. Adv. Neural Inf. Process. Syst.},
  pages={5998--6008},
  year={2017}
}

@article{gu2023mamba,
  title={Mamba: Linear-time sequence modeling with selective state spaces},
  author={Gu, Albert and Dao, Tri},
  journal={arXiv:2312.00752},
  year={2023}
}

@inproceedings{qi2017pointnet,
  title={Point{N}et: Deep learning on point sets for 3{D} classification and segmentation},
  author={Qi, Charles R and Su, Hao and Mo, Kaichun and Guibas, Leonidas J},
  booktitle={Proc. IEEE/CVF Conf. Comput. Vis. Pattern Recognit. (CVPR)},
  pages={652--660},
  year={2017}
}

@inproceedings{qi2017pointnet++,
  title={Pointnet++: Deep hierarchical feature learning on point sets in a metric space},
  author={Qi, Charles Ruizhongtai and Yi, Li and Su, Hao and Guibas, Leonidas J},
  booktitle={Proc. Adv. Neural Inf. Process. Syst.},
  volume={30},
  year={2017}
}

@article{wang2019dgcnn,
  title={Dynamic graph cnn for learning on point clouds},
  author={Wang, Yue and Sun, Yongbin and Liu, Ziwei and Sarma, Sanjay E and Bronstein, Michael M and Solomon, Justin M},
  journal={ACM Trans. Graph.},  
  volume={38},
  number={5},
  pages={1--12},
  year={2019},
  publisher={Acm New York, NY, USA}
}

@article{guo2021pct,
  title={P{C}{T}: Point cloud transformer},
  author={Guo, Meng-Hao and Cai, Jun-Xiong and Liu, Zheng-Ning and Mu, Tai-Jiang and Martin, Ralph R and Hu, Shi-Min},
  journal={Comput. Vis. Media},  
  volume={7},
  pages={187--199},
  year={2021},
  publisher={Springer}
}

@inproceedings{wang2018lsgcn,
  title={Local spectral graph convolution for point set feature learning},
  author={Wang, Chu and Samari, Babak and Siddiqi, Kaleem},
  booktitle={Proc. Eur. Conf. Comput. Vis.},
  pages={52--66},
  year={2018}
}

@inproceedings{huang2021awt-net,
  title={Adaptive wavelet transformer network for 3{D} shape representation learning},
  author={Huang, Hao and Fang, Yi},
  booktitle={Proc. Int. Conf. Learn. Represent.},
  pages={21471--21486},
  year={2021}
}

@article{shuman2013graph-signal-proc,
  title={The emerging field of signal processing on graphs: Extending high-dimensional data analysis to networks and other irregular domains},
  author={Shuman, David I and Narang, Sunil K and Frossard, Pascal and Ortega, Antonio and Vandergheynst, Pierre},
  journal={IEEE Signal Process. Mag.},
  volume={30},
  number={3},
  pages={83--98},
  year={2013},
  publisher={IEEE}
}

@article{hammond2011graph-wavelets,
  title={Wavelets on graphs via spectral graph theory},
  author={Hammond, David K and Vandergheynst, Pierre and Gribonval, R{\'e}mi},
  journal={Appl. Comput. Harmon. Anal.},
  volume={30},
  number={2},
  pages={129--150},
  year={2011},
  publisher={Elsevier}
}

@inproceedings{shen2023pointcmp,
  title={Pointcmp: Contrastive mask prediction for self-supervised learning on point cloud videos},
  author={Shen, Zhiqiang and Sheng, Xiaoxiao and Wang, Longguang and Guo, Yulan and Liu, Qiong and Zhou, Xi},
  booktitle={Proc. IEEE/CVF Conf. Comput. Vis. Pattern Recognit. (CVPR)},
  pages={1212--1222},
  year={2023}
}

@inproceedings{sheng2023pointcpsc,
  title={Point contrastive prediction with semantic clustering for self-supervised learning on point cloud videos},
  author={Sheng, Xiaoxiao and Shen, Zhiqiang and Xiao, Gang and Wang, Longguang and Guo, Yulan and Fan, Hehe},
  booktitle={Proc. IEEE/CVF Int. Conf. Comput. Vis.},
  booktitle={Proc. IEEE/CVF Int. Conf. Comput. Vis.},
  
  pages={16515--16524},
  year={2023}
}

@inproceedings{chen2024kan-hyperpointNet,
  title={KAN-{H}yperpoint{N}et for point cloud sequence-based 3{D} human action recognition},
  author={Chen, Zhaoyu and Li, Xing and Huang, Qian and Geng, Qiang and Yang, Tianjin and Han, Shihao},
  booktitle={Proc. IEEE Int. Conf. Acoust., Speech Signal Process. (ICASSP)},
  pages={1--5},
  year={2025},
  organization={IEEE}
}

@article{he2024prenet,
  title={P{R}{E}Net: A Plane-Fit Redundancy Encoding Point Cloud Sequence Network for Real-Time 3{D} Action Recognition},
  author={He, Shenglin and Qu, Xiaoyang and Wan, Jiguang and Li, Guokuan and Xie, Changsheng and Wang, Jianzong},
  journal={arXiv:2405.06929},
  year={2024}
}

@article{li2022SequentialPointNet,
  title={Real-{T}ime 3-{D} Human Action Recognition Based on Hyperpoint Sequence},
  author={Li, Xing and Huang, Qian and Wang, Zhijian and Yang, Tianjin and Hou, Zhenjie and Miao, Zhuang},
  journal={IEEE Trans. on Ind. Informat.},  
  volume={19},
  number={8},
  pages={8933--8942},
  year={2022},
  publisher={IEEE}
}

@inproceedings{li2010msr-action3d,
  title={Action recognition based on a bag of 3{D} points},
  author={Li, Wanqing and Zhang, Zhengyou and Liu, Zicheng},
  booktitle={Proc. IEEE Comput. Soc. Conf. Comput. Vis. Pattern Recognit. Workshops},  
  pages={9--14},
  year={2010},
  organization={IEEE}
}

@inproceedings{shahroudy2016ntu-rgbd,
  title={Ntu {R}{G}{B}+{D}: A large scale dataset for 3{D} human activity analysis},
  author={Shahroudy, Amir and Liu, Jun and Ng, Tian-Tsong and Wang, Gang},
  booktitle={Proc. IEEE/CVF Conf. Comput. Vis. Pattern Recognit. (CVPR)},
  pages={1010--1019},
  year={2016}
}

@article{liu2019ntu120,
  title={Ntu {R}{G}{B}+{D} 120: A large-scale benchmark for 3{D} human activity understanding},
  author={Liu, Jun and Shahroudy, Amir and Perez, Mauricio and Wang, Gang and Duan, Ling-Yu and Kot, Alex C},
  journal={IEEE Trans. Pattern Anal. Mach. Intell.},
  volume={42},
  number={10},
  pages={2684--2701},
  year={2019},
  publisher={IEEE}
}

@article{du2025prg-net,
  title={P{R}{G}-{N}et: Point Relationship-Guided Network for 3D human action recognition},
  author={Du, Yao and Hou, Zhenjie and Lin, En and Li, Xing and Liang, Jiuzhen and Zhou, Xinwen},
  journal={Neurocomputing},
  volume={635},
  pages={130015},
  year={2025},
  publisher={Elsevier}
}

@inproceedings{chang2024skele-wave-contrastive,
  title={Wavelet-decoupling contrastive enhancement network for fine-grained skeleton-based action recognition},
  author={Chang, Haochen and Chen, Jing and Li, Yilin and Chen, Jixiang and Zhang, Xiaofeng},
  booktitle={Proc. IEEE Int. Conf. Acoust., Speech Signal Process. (ICASSP)},
  pages={4060--4064},
  year={2024},
  organization={IEEE}
}

@inproceedings{qin2022strengthening,
  title={Strengthening skeletal action recognizers via leveraging temporal patterns},
  author={Qin, Zhenyue and Ji, Pan and Kim, Dongwoo and Liu, Yang and Anwar, Saeed and Gedeon, Tom},
  booktitle={Proc. Eur. Conf. Comput. Vis.},
  pages={577--593},
  year={2022}
}

@article{li2020prototypical,
  title={Prototypical contrastive learning of unsupervised representations},
  author={Li, Junnan and Zhou, Pan and Xiong, Caiming and Hoi, Steven CH},
  journal={arXiv:2005.04966},
  year={2020}
}

@inproceedings{selvaraju2017gradcam,
  title={Grad-{C}{A}{M}: Visual explanations from deep networks via gradient-based localization},
  author={Selvaraju, Ramprasaath R and Cogswell, Michael and Das, Abhishek and Vedantam, Ramakrishna and Parikh, Devi and Batra, Dhruv},
  booktitle={Proc. IEEE Int. Conf. Comput. Vis. (ICCV)},
  pages={618--626},
  year={2017}
}

@inproceedings{su2025ri,
  title={R{I}-{M}{A}{E}: Rotation-Invariant Masked AutoEncoders for Self-Supervised Point Cloud Representation Learning},
  author={Su, Kunming and Wu, Qiuxia and Cai, Panpan and Zhu, Xiaogang and Lu, Xuequan and Wang, Zhiyong and Hu, Kun},
  booktitle={Proc. AAAI Conf. on Artif. Intell.},
  volume={39},
  number={7},
  pages={7015--7023},
  year={2025}
}

@inproceedings{wu2025dc,
  title={D{C}-{P}{C}{N}: Point Cloud Completion Network with Dual-Codebook Guided Quantization},
  author={Wu, Qiuxia and Huang, Haiyang and Su, Kunming and Wang, Zhiyong and Hu, Kun},
  booktitle={Proc. AAAI Conf. on Artif. Intell.},
  volume={39},
  number={8},
  pages={8441--8449},
  year={2025}
}

@inproceedings{wu2024Point-transformer-v3,
  title={Point transformer v3: Simpler faster stronger},
  author={Wu, Xiaoyang and Jiang, Li and Wang, Peng-Shuai and Liu, Zhijian and Liu, Xihui and Qiao, Yu and Ouyang, Wanli and He, Tong and Zhao, Hengshuang},
  booktitle={Proc. IEEE/CVF Conf. Comput. Vis. Pattern Recognit. (CVPR)},
  pages={4840--4851},
  year={2024}
}

@inproceedings{ma2022rethinking,
  title={Rethinking network design and local geometry in point cloud: A simple residual MLP framework},
  author={Ma, Xu and Qin, Can and You, Haoxuan and Ran, Haoxi and Fu, Yun},
  booktitle={Proc. Int. Conf. Learn. Represent.},
  pages={5019--5033},
  year={2022}
}

@inproceedings{zhang2024safdnet,
  title={Safdnet: A simple and effective network for fully sparse 3D object detection},
  author={Zhang, Gang and Chen, Junnan and Gao, Guohuan and Li, Jianmin and Liu, Si and Hu, Xiaolin},
  booktitle={Proc. IEEE/CVF Conf. Comput. Vis. Pattern Recognit. (CVPR)},
  pages={14477--14486},
  year={2024}
}

@inproceedings{zhang2024voxel-mamba,
  title={Voxel mamba: Group-free state space models for point cloud based 3D object detection},
  author={Zhang, Guowen and Fan, Lue and He, Chenhang and Lei, Zhen and ZHANG, ZHAO-XIANG and Zhang, Lei},
  booktitle={Proc. Adv. Neural Inf. Process. Syst.},
  volume={37},
  pages={81489--81509},
  year={2024}
}

@inproceedings{zhu2024no-time-to-train,
  title={No time to train: Empowering non-parametric networks for few-shot 3D scene segmentation},
  author={Zhu, Xiangyang and Zhang, Renrui and He, Bowei and Guo, Ziyu and Liu, Jiaming and Xiao, Han and Fu, Chaoyou and Dong, Hao and Gao, Peng},
  booktitle={Proc. IEEE/CVF Conf. Comput. Vis. Pattern Recognit. (CVPR)},
  pages={3838--3847},
  year={2024}
}

@inproceedings{zhao2024unimix,
  title={Unimix: Towards domain adaptive and generalizable lidar semantic segmentation in adverse weather},
  author={Zhao, Haimei and Zhang, Jing and Chen, Zhuo and Zhao, Shanshan and Tao, Dacheng},
  booktitle={Proc. IEEE/CVF Conf. Comput. Vis. Pattern Recognit. (CVPR)},
  pages={14781--14791},
  year={2024}
}

@inproceedings{wang2024semantic-complete-4d,
  title={Semantic complete scene forecasting from a 4d dynamic point cloud sequence},
  author={Wang, Zifan and Ye, Zhuorui and Wu, Haoran and Chen, Junyu and Yi, Li},
  booktitle={Proc. AAAI Conf. on Artif. Intell.},
  volume={38},
  number={6},
  pages={5867--5875},
  year={2024}
}

@inproceedings{jing2024x4d-sceneformer,
  title={X4d-sceneformer: Enhanced scene understanding on 4D point cloud videos through cross-modal knowledge transfer},
  author={Jing, Linglin and Xue, Ying and Yan, Xu and Zheng, Chaoda and Wang, Dong and Zhang, Ruimao and Wang, Zhigang and Fang, Hui and Zhao, Bin and Li, Zhen},
  booktitle={Proc. AAAI Conf. on Artif. Intell.},
  volume={38},
  number={3},
  pages={2670--2678},
  year={2024}
}

@inproceedings{liu2023leaf,
  title={LeaF: learning frames for 4D point cloud sequence understanding},
  author={Liu, Yunze and Chen, Junyu and Zhang, Zekai and Huang, Jingwei and Yi, Li},
  booktitle={Proc. IEEE/CVF Int. Conf. Comput. Vis.},
  pages={604--613},
  year={2023}
}

@article{xiao2023unsupervised-pc-completion,
  title={Distinguishing and matching-aware unsupervised point cloud completion},
  author={Xiao, Haihong and Li, Yuqiong and Kang, Wenxiong and Wu, Qiuxia},
  journal={IEEE Trans. Circuits Syst. Video Technol.},
  volume={33},
  number={9},
  pages={5160--5173},
  year={2023},
  publisher={IEEE}
}

@article{liu2021geometrymotion,
  title={Geometry{M}otion-{N}et: A strong two-stream baseline for 3D action recognition},
  author={Liu, Jiaheng and Xu, Dong},
  journal={IEEE Trans. Circuits Syst. Video Technol.},
  volume={31},
  number={12},
  pages={4711--4721},
  year={2021},
  publisher={IEEE}
}

@article{tong2024one-shot-pc-action,
  title={You Will Never Walk Alone: One-Shot 3D Action Recognition With Point Cloud Sequence},
  author={Tong, Xingyu and Xiao, Yang and Tan, Bo and Yang, Jianyu and Cao, Zhiguo and Zhou, Joey Tianyi and Yuan, Junsong},
  journal={IEEE Trans. Circuits Syst. Video Technol.},
  volume={34},
  number={11},
  pages={11464--11477},
  year={2024},
  publisher={IEEE}
}

@article{song2024freq-image-cls,
  title={Interactive spectral-spatial transformer for hyperspectral image classification},
  author={Song, Liangliang and Feng, Zhixi and Yang, Shuyuan and Zhang, Xinyu and Jiao, Licheng},
  journal={IEEE Trans. Circuits Syst. Video Technol.},
  volume={34},
  number={9},
  pages={8589--8601},
  year={2024},
  publisher={IEEE}
}

@article{zhang2024freq-face-forgery-detection,
  title={Temporal diversified self-contrastive learning for generalized face forgery detection},
  author={Zhang, Rongchuan and He, Peisong and Li, Haoliang and Wang, Shiqi and Cao, Yun},
  journal={IEEE Trans. Circuits Syst. Video Technol.},
  year={2024},
  publisher={IEEE}
}

@article{wang2024fdnet,
  title={Fdnet: Frequency decomposition network for learned image compression},
  author={Wang, Jian and Ling, Qiang},
  journal={IEEE Trans. Circuits Syst. Video Technol.},
  volume={34},
  number={11},
  pages={11241--11255},
  year={2024},
  publisher={IEEE}
}

@article{wave-base-haar1910,
  author  = {Alfred Haar},
  title   = {Zur Theorie der orthogonalen Funktionensysteme},
  journal = {Mathematische Annalen},
  volume  = {69},
  pages   = {331--371},
  year    = {1910},
  doi     = {10.1007/BF01456326}
}

@article{wave-base-db1988,
  author  = {\vspace{0mm}Ingrid Daubechies},
  title   = {Orthonormal Bases of Compactly Supported Wavelets},
  journal = {Commun. Pure Appl. Math.},  
  volume  = {41},
  number  = {7},
  pages   = {909--996},
  year    = {1988},
  doi     = {10.1002/cpa.3160410705}
}

@book{wave-base-sym1992,
  author    = {Ingrid Daubechies},
  title     = {Ten Lectures on Wavelets},
  publisher = {Society for Industrial and Applied Mathematics},
  address   = {Philadelphia, PA},
  year      = {1992},
  series    = {CBMS-NSF Regional Conference Series in Applied Mathematics},
  volume    = {61},
  doi       = {10.1137/1.9781611970104}
}

@article{yan2020group-action-recg,
  title={Hi{G}{C}{I}{N}: Hierarchical graph-based cross inference network for group activity recognition},
  author={Yan, Rui and Xie, Lingxi and Tang, Jinhui and Shu, Xiangbo and Tian, Qi},
  journal={IEEE Trans. Pattern Anal. Mach. Intell.},
  volume={45},
  number={6},
  pages={6955--6968},
  year={2020},
  publisher={IEEE}
}

@article{yan2023compsi-action-recg,
  title={Progressive instance-aware feature learning for compositional action recognition},
  author={Yan, Rui and Xie, Lingxi and Shu, Xiangbo and Zhang, Liyan and Tang, Jinhui},
  journal={IEEE Trans. Pattern Anal. Mach. Intell.},
  volume={45},
  number={8},
  pages={10317--10330},
  year={2023},
  publisher={IEEE}
}

@article{qu2026few-shot-llm-st,
  title={Spatio-temporal Decoupled Knowledge Compensator for Few-Shot Action Recognition},
  author={Qu, Hongyu and Shu, Xiangbo and Yan, Rui and Gao, Hailiang and Wang, Wenguan and Tang, Jinhui},
  journal={IEEE Trans. Pattern Anal. Mach. Intell.},
  year={2026},
  publisher={IEEE}
}

@article{qu2025few-shot-mvp,
  title={M{V}{P}-{S}hot: Multi-velocity progressive-alignment framework for few-shot action recognition},
  author={Qu, Hongyu and Yan, Rui and Shu, Xiangbo and Gao, Hailiang and Huang, Peng and Xie, Guo-Sen},
  journal={IEEE Trans. Multimedia},
  year={2025},
  publisher={IEEE}
}

@article{xing2025locality,
  title={Locality-aware cross-modal correspondence learning for dense audio-visual events detection},
  author={Xing, Ling and Qu, Hongyu and Yan, Rui and Shu, Xiangbo and Tang, Jinhui},
  journal={IEEE Trans. Circuits Syst. Video Technol.},
  year={2025},
  publisher={IEEE}
}

@inproceedings{chen2026av-ssan,
  title={A{V}-{S}{S}{A}{N}: Audio-Visual Selective DOA Estimation Through Explicit Multi-Band Semantic-Spatial Alignment},
  author={Chen, Yu and Zhu, Hongxu and Wang, Jiadong and Chen, Kainan and Qian, Xinyuan},
  booktitle={Proc. AAAI Conf. on Artif. Intell.},
  volume={40},
  number={25},
  pages={20409--20417},
  year={2026}
}

@inproceedings{yang2022delving-freq,
  title={Delving into the frequency: Temporally consistent human motion transfer in the fourier space},
  author={Yang, Guang and Liu, Wu and Liu, Xinchen and Gu, Xiaoyan and Cao, Juan and Li, Jintao},
  booktitle={Proc. ACM Int. Conf. on Multimedia},
  pages={1156--1166},
  year={2022}
}

@inproceedings{zheng2022gait-reg-wild,
  title={Gait recognition in the wild with dense 3d representations and a benchmark},
  author={Zheng, Jinkai and Liu, Xinchen and Liu, Wu and He, Lingxiao and Yan, Chenggang and Mei, Tao},
  booktitle={Proc. IEEE/CVF Conf. Comput. Vis. Pattern Recognit. (CVPR)},
  pages={20228--20237},
  year={2022}
}

@article{dou2026dna,
  title={D{N}{A}: Uncovering Universal Latent Forgery Knowledge},
  author={Dou, Jingtong and Shi, Chuancheng and Wang, Yemin and Guo, Shiming and Yi, Anqi and Wu, Wenhua and Zhang, Li and Shen, Fei and Chua, Tat-Seng},
  journal={arXiv:2601.22515},
  year={2026}
}

@article{dou2026beyond,
  title={Beyond Surface Artifacts: Capturing Shared Latent Forgery Knowledge Across Modalities},
  author={Dou, Jingtong and Shi, Chuancheng and Wang, Jian and Shen, Fei and Wang, Zhiyong and Chua, Tat-Seng},
  journal={arXiv:2604.07763},
  year={2026}
}

\end{document}